\documentclass{article}

\PassOptionsToPackage{numbers}{natbib}
\usepackage[main, final]{neurips_2026}

\usepackage[utf8]{inputenc} 
\usepackage[T1]{fontenc}    
\usepackage{hyperref}       
\usepackage{url}            
\usepackage{booktabs}       
\usepackage{amsfonts}       
\usepackage{nicefrac}       
\usepackage{microtype}      
\usepackage[table]{xcolor}
\usepackage{array}
\usepackage{makecell}
\usepackage{graphicx}
\usepackage{xspace}
\usepackage{multirow}
\usepackage{siunitx}
\usepackage{tabularx}
\usepackage{xltabular}
\usepackage{amsmath}
\usepackage{wrapfig}
\usepackage{threeparttable}
\usepackage{caption}
\usepackage{makecell}
\usepackage{listing}
\usepackage{minted}
\usepackage{subcaption}
\usepackage{longtable}
\usepackage{float}
\usepackage{algorithm} 
\usepackage{algpseudocode}
\usepackage{enumitem}
\usepackage{titletoc}
\usepackage{booktabs}
\usepackage{adjustbox}
\usepackage{graphicx}
\usepackage{threeparttable}
\usepackage[table]{xcolor}
\usepackage{fontawesome5}
\newcommand{\avgplain}[2]{%
  #1{\scriptsize\textcolor{black}{\,(\textbf{#2})}}%
}

\newcommand{\scoreplain}[2]{%
  #1{\scriptsize\textcolor{black}{\,(\textbf{#2})}}%
}

\usepackage[most]{tcolorbox}
\usepackage{xcolor}
\usepackage{listings}

\definecolor{promptblue}{HTML}{5F759D}
\definecolor{promptbg}{HTML}{F8F9FC}
\definecolor{promptpink}{HTML}{D81B60}
\definecolor{softblue}{HTML}{3485BA}

\lstdefinestyle{promptstyle}{
    basicstyle=\small\rmfamily,
    breaklines=true,
    columns=fullflexible,
    keepspaces=true,
    showstringspaces=false,
    upquote=true,
    literate=
      {\{question\}}{\textcolor{promptpink}{\{question\}}}{10}
      {\{answer\}}{\textcolor{promptpink}{\{answer\}}}{8}
      {—}{{\textemdash}}1
      {–}{{\textendash}}1
      {‘}{{\textquoteleft}}1
      {’}{{\textquoteright}}1
      {“}{{\textquotedblleft}}1
      {”}{{\textquotedblright}}1
      {…}{{\textellipsis}}1
}

\lstdefinestyle{promptstyle}{
    basicstyle=\small\rmfamily,
    breaklines=true,
    columns=fullflexible,
    keepspaces=true,
    showstringspaces=false,
    upquote=true,
    literate=
      {\{question\}}{\textcolor{promptpink}{\{question\}}}{10}
      {\{answer\}}{\textcolor{promptpink}{\{answer\}}}{8}
}

\newtcblisting{promptbox}[1]{
    listing only,
    breakable,
    enhanced,
    colback=promptbg,
    colframe=promptblue!80!black,
    colbacktitle=promptblue,
    coltitle=white,
    title=\textbf{#1},
    fonttitle=\bfseries,
    boxrule=0.6pt,
    arc=1pt,
    left=2mm,
    right=2mm,
    top=1mm,
    bottom=1mm,
    listing options={style=promptstyle}
}

\definecolor{citecolor}{HTML}{0071bc}
\definecolor{cadmiumgreen}{rgb}{0.0, 0.42, 0.24}
\definecolor{cornellred}{rgb}{0.7, 0.11, 0.11}
\hypersetup{
  linkcolor = cornellred,
  citecolor  = citecolor,
  colorlinks = true,
}
\definecolor{Gray2}{gray}{0.7}
\def \name{\textsc{HuB-J}\xspace}

\title{Auditing the Judge: Human-Derived Bias Discovery, Quantification, and Mitigation in LLM Judges}

\author{%
  Hamin Koo$^{1}$\thanks{Equal contribution.}
  \quad
  Chanjoo Jung$^{1}$\footnotemark[1]
  \quad
  Fangzhao Wu$^{2}$
  \quad
  Jaehyung Kim$^{1}$
  \\[0.5em]
  $^{1}$Yonsei University
  \qquad
  $^{2}$Microsoft Research Asia
  \\[0.3em]
  \texttt{\{hamin2065, chanjoo0427\}@yonsei.ac.kr}
}

\begin{document}

\maketitle
\vspace{-1.5em}
\begin{center}
\begin{tabular}{@{}l@{}}
\faGlobe\enspace
\textbf{Project page:}
\href{https://hamin2065.github.io/HUB-J_project_page/}
{\textcolor{softblue}{\texttt{github.io/HUB-J\_project\_page/}}}\\
\faGithub\enspace
\textbf{Code:}
\href{https://github.com/hamin2065/HUB-J}
{\textcolor{softblue}{\texttt{github.com/hamin2065/HUB-J}}}
\end{tabular}
\end{center}
\vspace{1.2em}

\begin{abstract}
While Large Language Models (LLMs) are increasingly deployed as automated evaluators in evaluation and training pipelines, their judgements are often affected by systematic biases that conflict with human preferences. 
Although prior work has identified various biases and proposed methods for detecting and mitigating them, less attention has been paid to using human preferences to inform bias identification and mitigation.
Moreover, these approaches rely heavily on pre-discovered bias lists, overlook bias strength, and depend on costly interventions. 
In this work, we propose \name{}, an integrated framework based on human-derived bias factors that detects, quantifies, and mitigates biases in LLM-as-a-Judge systems. 
Our approach leverages human–LLM judgement disagreement cases to automatically discover interpretable bias factors, and utilizes agreement cases to quantify bias strength through controlled input modifications and resulting shifts in model decisions. 
Finally, building on these quantified biases, we introduce a lightweight, training-free regression-based mitigation strategy that estimates the influence of each bias factor on model judgments and corrects bias-influenced decisions by removing its estimated effect. Empirical results show that \name{} uncovers both known and novel bias factors, reveals meaningful differences in model susceptibility, and consistently reduces bias-driven decision flips while generalizing across models.

\end{abstract}

\section{Introduction}

With recent advancements in Large Language Models (LLMs), LLM-based evaluation (often referred to as \textit{LLM-as-a-Judge}) has been increasingly adopted in both evaluation and training pipelines, making their reliability an important problem~\citep{zheng2023judging, ye2025learning, liu2023g,ong2025towards, seo2026fast}.
One major concern is systematic evaluation biases~\citep{wang2024large, zeng2024evaluating, dubois2024length} in LLM judges, where LLM judges prefer one response over another based on factors that do not reflect genuine response quality or human evaluators' preferences.
For example, \citep{zheng2023judging} identified position and verbosity biases, where LLM judges favor responses based on their presentation order or length rather than the actual content of the answer. 
However, manually discovering such biases remains challenging, as they are often implicit and costly to annotate at scale. 
This has motivated recent studies on automatic bias discovery and analysis with LLMs\citep{lai2026biasscope,ye2025justice}. 

\vspace{1em}
However, prior approaches face fundamental limitations.
First, in \textit{bias identification}, the exploration of new candidates relies heavily on pre-discovered bias lists or LLM reasoning, which inherently constrains the search space; 
moreover, this identification is conducted in an instance-wise and model-specific manner, which limits the generalizability of the discovered biases \citep{lai2026biasscope}.
Second, in \textit{bias quantification}, prior studies largely treat bias as a binary phenomenon whether a bias flips the judge's decision, offering limited insight into how strongly each bias manifests in a given judge \citep{ye2025justice}. 
Third, in \textit{bias mitigation}, many existing approaches~\citep{sant2024power, Yang2026FairJudgeAA} depend on model-specific interventions, such as additional training or prompt engineering, which incur substantial training or engineering costs.
Overall, these limitations call for an integrated framework that discovers, quantifies, and mitigates biases in a manner informed by human evaluation preferences.

\vspace{1em}
\textbf{Contribution.}
In this work, we propose \name{}, a unified framework to 1) automatically discover \textbf{Hu}man-derived \textbf{B}ias factors in LLM \textbf{J}udges, 2) quantify their model-specific effects, and 3) perform lightweight debiasing.
Our key idea is to derive bias factors from human-LLM judgement disagreements by using publicly available human preference datasets: disagreement cases between human labels and LLM judgments are used to identify biases that drive human--LLM judgment gaps, while agreement cases are used to quantify how strongly each bias can affect judgments. 
Specifically, \name{} first identifies bias factors from cases where human and model judgments disagree by extracting instance-level features and grouping them into interpretable factors. 
This allows us to discover biases that emerge from actual disagreement without relying on a fixed list of known biases.
Second, \name{} quantifies a bias factor using cases where human and model judgments agree; we define a 1--5 Likert scale for bias intensity and modify the rejected response (\textit{i.e.}, the one neither the human nor the model selected) to reflect different levels of the factor.
We then measure whether the judge's decision flips in favor of the modified response, and use this flip rate to estimate each model's susceptibility to the factor and derive a model-specific bias profile.
Finally, we propose a lightweight regression-based approach to estimate bias effects and recover counterfactual judgments in their absence.

\vspace{1em}
The empirical results validate the effectiveness of \name{} across all three stages of the framework. 
In bias identification, \name{} successfully discovers bias factors that are strongly supported by disagreement signals, while also revealing novel bias factors underlying judge misalignment.
For bias quantification, \name{} reveals model-specific bias preference profiles, showing that different LLM judges are susceptible to different factors and effective intensity levels.
For instance, \textit{Technical Jargon}--a newly discovered, human-interpretable factor not covered by the baseline bias sets--shows that different judges respond to it very differently. 
{Also, an extended experiment with 20 state-of-the-art LLMs further shows that these bias patterns persist across a broader model pool, suggesting that the discovered factors capture generalizable judge-specific vulnerabilities.}
Finally, our proposed bias mitigation method reduces bias-induced corruption in judgments, correcting an average of 31.8\% of bias-driven preference deviations.
In fact, it remains robust across five judges spanning diverse model families and scales.
Overall, these results demonstrate that \name provides a unified and effective framework for discovering, quantifying, and mitigating biases in LLM-as-a-Judge.

\begin{figure}
    \centering
    \footnotesize
    \includegraphics[width=1.0\linewidth]{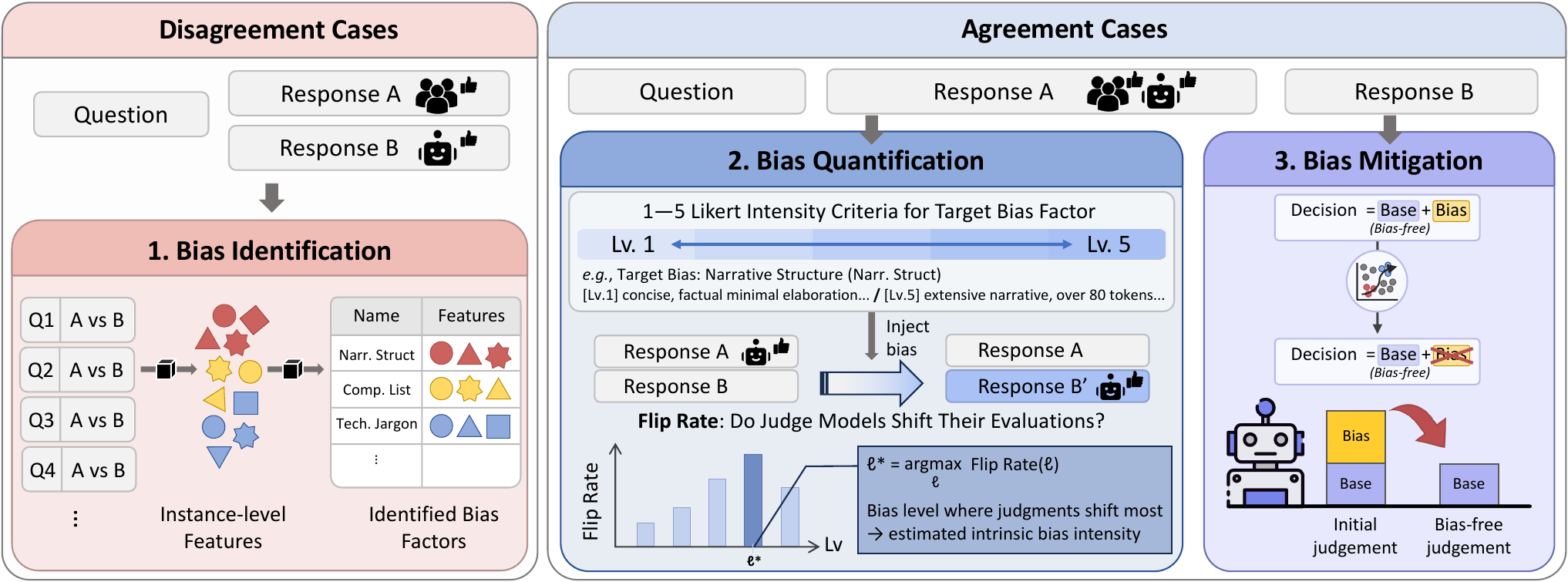}
    \caption{
\textbf{Overview of \name{}}, a unified framework for discovering, quantifying, and mitigating human-derived bias in LLM-as-a-judge systems. 
(1) \textit{Bias identification} uses human–LLM disagreement cases to extract and cluster instance-level features into bias factors. 
(2) \textit{Bias quantification} uses agreement cases by injecting each factor into rejected responses across 1--5 Likert levels, defining effective intensity as the point of strongest model decision shift. 
(3) \textit{Bias mitigation} decomposes judgments into bias-free and bias-attributable parts, enabling bias estimation and correction.
}
\label{fig:main_figure}
\vspace{-5mm}
\end{figure}

\section{Related Work}

\noindent\textbf{LLM-as-a-Judge.}
LLM-as-a-Judge has emerged as a scalable alternative to human evaluation in open-ended generation, especially in cases where reference-based metrics fail to reflect human preferences~\citep{koo2025emcee}. Results from MT-Bench and Chatbot Arena show that strong LLMs can operate effectively as pairwise evaluators, motivating their use in model comparison and benchmark development~\citep{zheng2023judging}. More recent work has developed evaluator models, automatic evaluation protocols, and meta-evaluation benchmarks, including G-Eval, PandaLM, Prometheus, Length-Controlled AlpacaEval, LLMBar, and JudgeBench~\citep{liu2023g,wang2024pandalm,kim2024prometheus,dubois2024length,zeng2024evaluating,tan2025judgebench}. 
Beyond evaluation, preference judgements are also important to alignment pipelines such as reward modeling and reinforcement learning from human or AI feedback~\citep{ouyang2022training,bai2022constitutional,lee2024rlaif}. 
Since these signals influence both model selection and optimization, their reliability is becoming increasingly important in practice. \looseness-1

\vspace{1mm}
\noindent\textbf{Bias in LLM-as-a-Judge.}
Prior work shows that LLM judges can be influenced by superficial factors such as response order, length, and model identity rather than underlying response quality, leading to position, verbosity, self-enhancement, and related biases~\citep{raina-etal-2024-llm, zheng2023judging,wang2024large, chen2024humans, panickssery2024llm, jeong2025comparative}. 
Subsequent studies quantify pre-discovered bias dimensions, evaluate judge robustness with challenging benchmarks, and explore automatic bias discovery or debiased judge training~\citep{ye2025justice,tan2025judgebench,zeng2024evaluating,lai2026biasscope,Yang2026FairJudgeAA}. 
Related work also shows that judge preferences can be reverse-engineered or exploited by optimizing against LLM-judge feedback~\citep{alazraki2025reverse}. 
At the same time, recent studies have explored automatic bias discovery in reward models, identification of implicit alignment objectives, and diagnosis and mitigation of stylistic biases in preference models \citep{wang2026automatically,chen2026discovering,bharadwaj2026flattery}.
However, existing approaches often rely on pre-discovered factors, validate biases through local perturbations, or treat bias as a binary flip event \citep{lai2026biasscope, ye2025justice}. 
In contrast, \name{} discovers human-derived bias factors from human–LLM disagreement, measures model-specific bias strength via controlled intensity scaling, and reduces their effects without retraining the judge.

\section{Method}

In this section, we present the formal procedure of \name{}. 
The full algorithmic details across all stages are provided in  Appendix~\ref{sec:app_algorithms}.

\subsection{Preliminaries} \label{sec:preliminary}

We consider a pairwise human preference dataset
\[
\mathcal{D}^{\mathrm{full}}
=
\{d_i=(x_i, r_i^A, r_i^B, y_i^H)\}_{i=1}^{N},
\]
where \(x_i\) is the prompt, \(r_i^A\) and \(r_i^B\) are the two candidate responses, and \(y_i^H \in \{A,B\}\) is the human preference label.
For each instance, we denote the human-preferred and human-rejected responses as
\[
r_i^+ = r_i^{y_i^H}, 
\qquad 
r_i^- = r_i^{\bar{y}_i^H},
\]
where \(\bar{y}_i^H\) is the unselected option. 
For a judge model \(m\) with prediction $y_i^m = m(x_i, r_i^A, r_i^B)$, we partition \(\mathcal{D}^{\mathrm{full}}\) into per-model agreement and disagreement datasets:
\[
\mathcal{D}^{\mathrm{agr}}_m
=
\{d_i \in \mathcal{D}^{\mathrm{full}} \mid y_i^m = y_i^H\},
\qquad
\mathcal{D}^{\mathrm{dis}}_m
=
\{d_i \in \mathcal{D}^{\mathrm{full}} \mid y_i^m \neq y_i^H\},
\]
so that, in $\mathcal{D}^{\mathrm{dis}}_m$, the judge $m$ selects the human-rejected response \(r_i^{-}\) over the human-preferred response \(r_i^{+}\) while, in $\mathcal{D}^{\mathrm{agr}}_m$, both the human and judge $m$ select $r_i^{+}$. 

This separation naturally suits the two stages of our framework. 
We use $\mathcal{D}^{\mathrm{dis}}_m$ for \textit{bias identification}, since cases where judge $m$ favors the human-rejected response can expose latent biases of the judge model that drive its decisions away from human preferences.
We use $\mathcal{D}^{\mathrm{agr}}_m$ for \textit{bias quantification} because these cases provide a controlled starting point: both humans and judge $m$ originally prefer $r_i^{+}$ over $r_i^{-}$.
We then inject a candidate bias factor into the originally rejected response $r_i^{-}$ and measure whether judge $m$ switches its preference toward the modified response.


Throughout the following sections, we denote LLM-based generation modules by \(G_{\bullet}\), where the subscript specifies the task performed by each module (\textit{e.g.}, $G_{\mathrm{extract}}$ for extracting interpretable features during the bias identification stage). The prompt for each module is provided in Appendix \ref{sec:app_prompts}.

\subsection{Bias Identification}  \label{sec:bias_identification}

To identify bias factors that generalize across models, we construct a consensus disagreement subset rather than relying on a single judge $m$, since using only one judge may introduce model-specific tendencies that limit generalization. 
Specifically, given a set of state-of-the-art LLM judges $\mathcal{M}=\{m_1, \dots, m_K\}$, we define a consensus disagreement subset as $\mathcal{D}^{\mathrm{dis}}_{\mathcal{M}}=\bigcap_{k=1}^{K} \mathcal{D}^{\mathrm{dis}}_{m_k}$. 
This focuses our analysis on cases where multiple LLM judges consistently disagree with human preferences, rather than being driven by model-specific effects. 
Inspired by \cite{kim2026rpm,wang-etal-2023-goal}, we subsequently conduct the following pipeline to identify and construct bias factors.\looseness-1

\textbf{Feature extraction}. For each instance \(d_i \in \mathcal{D}^{\mathrm{dis}}_{\mathcal{M}}\), we use a feature extraction module \(G_{\mathrm{extract}}\) to extract a set of features \(\mathcal{F}_i=\{f_{i,j}\}\), where each \(f_{i,j}\) captures a possible reason for the discrepancy between human and LLM judge preferences for the given response pair. 
Then, we aggregate these instance-level features into a global feature pool:
\[
\mathcal{F}
=
\bigcup_{d_i\in\mathcal{D}^{\mathrm{dis}}_{\mathcal{M}}}
\mathcal{F}_i.
\] 

\textbf{Factorization.} Using the extracted feature set $\mathcal{F}$, our goal is to construct a compact set of top-$K$ bias factors that best explain $\mathcal{F}$. 
To this end, we conduct the following processes inspired by RPM \citep{kim2026rpm}: 

\textbf{Stage 1: Candidate factor initialization.}
We first apply K-means clustering on the feature set $\mathcal{F}$ to group semantically similar features, where each cluster represents a potential latent bias region. From each cluster, we select the feature closest to the centroid as its representative. All representatives are passed to $G_{\mathrm{construct}}$, which generates a set of candidate bias factors
$
\mathcal{Z}^{\mathrm{cand}} = \{z_1, \ldots, z_M\},
$
together with textual descriptions $t_z$ for each factor. 
Then, to remove semantically redundant factors in $\mathcal{Z}^{\mathrm{cand}}$, we apply embedding-based similarity filtering. 
Each factor $z$ is represented by an embedding $e_z$, and we compute pairwise cosine similarity. We obtain a refined factor set $\tilde{\mathcal{Z}} \subseteq \mathcal{Z}^{\mathrm{cand}}$ such that redundant factors with similarity above threshold $\tau \in [0,1]$ are removed.

\textbf{Stage 2: Assignment and coverage estimation.}
Subsequently, we use an assignment module $G_{\mathrm{assign}}$ to map each feature to its most relevant factor or abstain if no factor sufficiently explains it:
\[
a = G_{\mathrm{assign}}(\mathcal{F}, \tilde{\mathcal{Z}}), \quad
a : \mathcal{F} \rightarrow \tilde{\mathcal{Z}} \cup \{\bot\}.
\] 
For each factor $z \in \tilde{\mathcal{Z}}$, we define its coverage set $\mathcal{C}(z) = \{ f \in \mathcal{F} \mid a(f) = z \}$ and the uncovered feature set $\mathcal{F}_{\mathrm{unc}} = \{ f \in \mathcal{F} \mid a(f) = \bot \}$.

\textbf{Stage 3: Greedy top-$K$ selection.}
Then, we construct the final factor set $\mathcal{Z}=\{z_1, \ldots, z_K\}$ by greedily selecting $K$ factors from $\tilde{\mathcal{Z}}$. At each step, we select:
\[
z^{*} =
\arg\max_{z \in \tilde{\mathcal{Z}} \setminus \mathcal{Z}}
\Big[
\alpha \cdot |\mathcal{C}(z) \setminus \mathcal{U}(\mathcal{Z})|
-
\lambda_{u} \cdot |\mathcal{F}_{\mathrm{unc}}|
\Big],
\]
where $\mathcal{U}(\mathcal{Z}) = \bigcup_{z \in \mathcal{Z}} \mathcal{C}(z)$ denotes the set of features already covered by the current factor set. $\alpha$ and $\lambda_u$ are hyperparameters that control the strength of each term in the objective.

The features not covered by the final selected factors are added to $\mathcal{F}_{\mathrm{unc}}$; 
then, we repeat Stages 1–3 using the current uncovered feature set $\mathcal{F}_{\mathrm{unc}}$, while retaining and updating the candidate factor pool across iterations. This process continues until the coverage ratio exceeds threshold $\rho$ or a maximum of $T$ iterations is reached.

\subsection{Bias Quantification} \label{sec:bias_quantification}

Given a bias factor \(z\) and its textual description \(t_z\), we quantify how strongly \(z\) affects a given judge model. 
To this end, we inject factor \(z\) into the originally rejected response in agreement cases and measure whether the judge flips toward the modified response.

\noindent\textbf{Bias injection with intensity criteria.} \label{tab:bias_criteria}
To control the strength of the bias factor, we first convert the factor description \(t_z\) into a five-level intensity criterion where lower intensity indicates weaker bias: 
\[
\mathcal{C}_z
=
G_{\mathrm{crit}}(z,t_z)
=
\{c_{z,\ell}\},
\]
where \(c_{z,\ell}\) specifies how the bias factor \(z\) should be expressed at intensity level \(\ell\in L\), $L=\{1,2,3,4,5\}$.
Each criterion pairs a description with factor-specific quantifiable anchors (\textit{e.g.}, token ranges). 
By grounding each level in observable response properties, these anchors allow us to inject the same factor at controlled strengths, ordered from the weakest to the strongest manifestation.

Then, for each agreement instance \(d_i\in\mathcal{D}^{agr}_m\) where both human 
and judge \(m\) select \(r_i^+\), we 
rewrite only the rejected response \(r_i^-\) according to the target factor $z$ and intensity level $\ell$:
\[
\tilde r^-_{i,z,\ell}
=
G_{\mathrm{inj}}(x_i,r_i^-,z,t_z,c_{z,\ell}).
\]

Here, \(G_{\mathrm{inj}}\) is a constrained rewriting module; an LLM is prompted to rewrite \(r_i^-\) conditioned on the target factor description \(t_z\) and intensity criterion
\(c_{z,\ell}\), but also preserve the task-relevant content, such as final answers and named entities.
This yields a controlled comparison pair \((r_i^+,\tilde r^-_{i,z,\ell})\), allowing us to measure whether the injected factor changes the judge's preference.

\paragraph{Flip rate and effective intensity.}
For each comparison between $r_i^+$ and $\tilde r^-_{i,z,\ell}$, we assign \(y^m_{i,z,\ell}=1\) when $m$ selects \(\tilde r^-_{i,z,\ell}\), \(0\) for selecting \(r_i^+\), and \(0.5\) for neutral cases such as ties or inconsistent decisions across orders.\footnote{To prevent position bias, we evaluate each comparison twice by swapping the order of the two responses.}
Then, for a set of agreement instances \(\mathcal{S}\subseteq \mathcal{D}^{\mathrm{agr}}_m\), we define its flip rate under factor \(z\) at intensity level \(\ell\) as follows:
\[
\mathrm{FR}_{m,z,\ell}(\mathcal{S})
=
\frac{1}{|\mathcal{S}|}
\sum_{d_i \in \mathcal{S}}
y_{i,z,\ell}^{m}.
\]
A higher flip rate indicates higher susceptibility to the injected degree of bias factor, as the judge more often reverses its original preference toward the modified rejected response.

To select the most effective intensity level, we split the agreement dataset 
\(\mathcal{D}^{\mathrm{agr}}_m\) into a calibration split 
\(\mathcal{D}^{\mathrm{cal}}_m\) and a held-out test split 
\(\mathcal{D}^{\mathrm{test}}_m\). 
For each judge model \(m\) and factor \(z\), we choose the intensity level that induces the highest flip rate on the calibration split:
\[
\ell_{m,z}^{*}
=
\arg\max_{\ell \in L}
\mathrm{FR}_{m,z,\ell}(\mathcal{D}^{\mathrm{cal}}_m).
\]
We call \(\ell^*_{m,z}\) the \textbf{model-specific effective intensity}. We then
fix this intensity and evaluate the flip rate on the held-out test split,
defining the \textbf{model-factor susceptibility score} as: 
\[
\rho_{m,z}
=
\mathrm{FR}_{m,z,\ell^*_{m,z}}(\mathcal{D}^{test}_m).
\]
Thus, \(\rho_{m,z}\) is the held-out flip rate of judge \(m\) for factor \(z\)
under its calibration-selected intensity. Higher values indicate that the
judge more often flips toward the bias-injected response.
Given discovered factors \(\mathcal{Z}=\{z_1,\ldots,z_K\}\), we collect the scores into a model-specific bias profile:  
\[
\boldsymbol{\rho}_m
=
[\rho_{m,z_1},\rho_{m,z_2},\ldots,\rho_{m,z_K}].
\]

\subsection{Bias Mitigation} \label{sec:bias_mitigation}

Let $y^{m} \in \{0,1\}$ denote judge $m$’s prediction on query $x \in \mathcal{D}$, \textit{i.e.}, whether the candidate response is preferred over the reference. 
Then, the win rate on dataset $\mathcal{D}$ is defined as $\mathrm{Win}_{m}(\mathcal{D}) = \mathbb{E}_{x \in \mathcal{D}}[y^{m}]$.  

Given a query $x$, we use a fixed scoring function $G_{\mathrm{score}}$ (see Section~\ref{tab:bias_criteria}) to assign each response a scalar intensity score $\ell \in \mathcal{L}$, reflecting the strength of the target bias factor. We compute scores for the judge-preferred and human-preferred responses, denoted as $\ell_{\text{judge}}$ and $\ell_{\text{human}}$, respectively, and define the bias intensity differential $\Delta$ as:
\[
\Delta = \ell_{\text{judge}} - \ell_{\text{human}}.
\] 
To quantify the contribution of the bias factor to judge predictions and mitigate its effect, we model each prediction as a logistic function of three components:
\[
y^{m} = \sigma\!\big(
\underbrace{\theta_m}_{\text{baseline}}
+ \underbrace{\phi_m\, \tanh(\Delta)}_{\text{bias sensitivity}}
+ \underbrace{\psi_m\, \gamma_x}_{\text{query effect}}
\big).
\]
Here, $\theta_m$ captures judge $m$'s baseline preference, $\phi_m$ measures sensitivity to the bias factor, $\gamma_x$ represents a shared query-level effect, and $\psi_m$ controls how strongly judge $m$ responds to this shared signal. Given this formulation, the model is not directly identifiable due to scale ambiguity between $\psi_m$ and $\gamma_x$. To address this, we estimate the parameters using a two-stage procedure, inspired by Length-Controlled AlpacaEval \cite{dubois2024length} (see Appendix~\ref{sec:app_mitigation} for details).

The estimated parameters enable computation of a counterfactual preference rate for judge $m$ in the absence of the bias factor. Specifically, removing the bias term and averaging over queries yields:
\[
\mathrm{BC}_m(\mathcal{D}) = \mathbb{E}_{x\in\mathcal{D}}\!\big[\sigma(\hat\theta_m + \hat\psi_m\, \hat\gamma_x)\big].
\]

We denote $\mathrm{BC}_m$ as the \textbf{bias-controlled win rate}, which removes the contribution of the bias term while preserving baseline and query-level effects. 
The difference $\Delta_m(\mathcal{D}) = \mathrm{Win}_{m}(\mathcal{D}) - \mathrm{BC}_m(\mathcal{D})$ quantifies the aggregate contribution of the bias factor to judge $m$'s observed win rate.

\section{Experiment} \label{sec:experiment}

\subsection{Setups}
\label{sec:setups}

\noindent\textbf{Benchmarks.}
We construct the datasets for the experiments by merging a collection of five open-ended, human-annotated preference benchmarks:
(1) \textit{MT-Bench}~\citep{zheng2023judging} for multi-turn chat evaluation,
(2) \textit{MultiPref}~\citep{miranda2025hybrid} for multi-annotated and multi-aspect preferences,
(3) \textit{WebGPT-Comparisons}~\citep{nakano2021webgpt} for long-form question-answering comparisons,
(4) \textit{HelpSteer2-Pref}~\citep{wang2024helpsteer2} for pairwise human preferences, and
(5) \textit{HelpSteer3}~\citep{wang2025helpsteer3} for human preference annotations over real-world LLM-use prompts.
This merged benchmark allows us to analyze when LLM judges systematically deviate from human preferences across diverse open-ended tasks, after splitting it into agreement and disagreement datasets (Section~\ref{sec:preliminary}).
For data with multiple human annotations, we retain only examples where at least 50\% of annotators select the same response, ensuring the reliability of human preference labels. 

\noindent\textbf{Models.} 
We use five representative LLMs as judge models: 
\textit{GPT-4o}~\citep{openai2024gpt4o}, 
\textit{Claude-Haiku-4.5}~\citep{anthropic2025haiku45}, 
\textit{Gemini-3-Flash-Preview}~\citep{google2025gemini3flash}, 
\textit{Qwen-Plus}~\citep{alibaba2026qwenplus}, and 
\textit{DeepSeek-V3.2}~\citep{deepseek2025v32}.
For brevity, we refer to these models as GPT, Haiku, Gemini, Qwen, and DeepSeek, respectively, throughout the paper.
To mitigate position bias, we prompt each judge twice per instance by swapping the order of the two candidate responses.
During data curation, we exclude tie cases but include them in flip-rate computation.
Unless otherwise specified, GPT-4o is employed as the generation module $G_{\bullet}$. 
All model judgments are generated with temperature set to $0$ for deterministic evaluation.

\setlength{\tabcolsep}{3pt} 
\begin{table}[t]
\centering
\footnotesize
\renewcommand{\arraystretch}{1.3}
\caption{\textbf{Identified top 5 bias factors of \name{}.} The abbreviations and definitions are provided in the table below. For clarity, we use these abbreviations throughout the paper.}
\vspace{0.05in}
\label{tab:our_bias_definitions}
\begin{tabularx}{\textwidth}{
@{\hspace{0.6em}}
>{\raggedright\arraybackslash}p{4.5cm}
@{\hspace{1.0em}}
X
@{\hspace{0.6em}}
}
\toprule
\textsc{Bias Name (Abbr.)} & \textsc{Definition} \\ 
\midrule
\textbf{Comprehensive List} \newline \textit{(Comp. List)} & A pattern where responses that provide longer or more comprehensive lists are favored, potentially mistaking quantity for relevance. \\ \addlinespace

\textbf{Professional Tone} \newline \textit{(Prof. Tone)} & The tendency to favor responses with a more professional or formal tone, interpreting them as more credible or authoritative. \\ \addlinespace

\textbf{Narrative Structure} \newline \textit{(Narr. Struct)} & A bias where responses structured as narratives or stories are favored, valuing the flow and progression of the content. \\ \addlinespace

\textbf{Implementation Focus} \newline \textit{(Impl. Focus)} & A bias where responses emphasizing implementation details and practical steps are preferred, potentially valuing them over broader strategic considerations. \\ \addlinespace

\textbf{Technical Jargon} \newline \textit{(Tech. Jargon)} & The tendency to favor responses that use technical jargon or specialized terms, perceiving them as more expert or authoritative. \\ 
\bottomrule
\end{tabularx}
\end{table}

\noindent\textbf{Baselines.} 
To demonstrate the effectiveness of \name{}, we compare against two types of baselines: 
(1) \textit{pre-discovered bias factors}, collected from prior literature~\citep{alazraki2025reverse, chen2024humans, ye2025justice,zheng2023judging}; and 
(2) \textit{BiasScope}~\citep{lai2026biasscope}, which identifies bias factors by leveraging LLM reasoning.
To align with the core methodology of the original paper, we prompt GPT-4o to analyze judge decision explanations for each response pair and extract candidate bias dimensions.
These candidates are then evaluated in a controlled validation setting, where we select the top factors based on how frequently they flip judges' decisions.

\noindent\textbf{Implementation details.} 
For bias identification, we use \textit{disagreement} dataset, where all five judge models select the same response while disagreeing with the human label, consisting of 1,830 examples split into 1,464 training and 366 test examples.
For the quantification and mitigation stages, we construct an \textit{agreement} dataset by filtering examples where all five judge models agree with the human annotation, using 100 examples for calibration and 500 examples for held-out evaluation.
Additional details on implementation, models, full prompts for all LLM-based modules, bias-intensity criteria, and baseline reproduction are provided in Appendices~\ref{sec:app_impl_details},~\ref{sec:app_prompts}, and Table~\ref{tab:bias_intensity_criteria}.

\subsection{Results on Bias Identification}

\textbf{\name{}'s identified bias factors.} \label{tab:bias_factors}
For comparison, we use five representative pre-discovered bias factors: \textit{Verbosity}, \textit{Authority}, \textit{Beauty}, \textit{Sentiment}, and \textit{Bandwagon}.
We also include five factors produced by \textsc{BiasScope}: \textit{Availability}, \textit{Complexity}, \textit{Illusion of Knowledge}, \textit{Length}, and \textit{Risk Perception}. 
Full descriptions of these baseline factors are provided in Appendix~\ref{sec:app_base_details}. 
On the other hand, \name{} extracts features directly from disagreement-case response pairs and aggregates them to construct bias factors. As presented in Table~\ref{tab:our_bias_definitions}, the resulting bias factor set of our method consists of \textbf{Comprehensive List}, \textbf{Professional Tone}, \textbf{Narrative Structure}, \textbf{Implementation Focus}, and \textbf{Technical Jargon}. 
Qualitatively, these factors are more directly reflected in response characteristics such as structure, level of implementation detail, and use of technical terminology. 
To quantitatively compare these factors, we introduce two metrics evaluating their effectiveness and novelty. 

\begin{table}[t]
\centering
\scriptsize
\setlength{\tabcolsep}{5pt}
\renewcommand{\arraystretch}{1.15}

\caption{\textbf{Comparison of bias identification results.}
Mean $\Delta$ is computed per response pair in human--LLM disagreement cases.
$^{*}$ indicates $p < 0.01$. 
Full p-values are reported in Table~\ref{tab:full_p_value}}.
\vspace{0.05in}
\label{tab:bias_identification_mean_delta}

\begin{tabular}{@{} 
l c 
l c 
l c @{}}
\toprule

\multicolumn{2}{c}{\textsc{Pre-discovered}} &
\multicolumn{2}{c}{\textsc{BiasScope~\citep{lai2026biasscope}}} &
\multicolumn{2}{c}{\textsc{Ours}} \\
\cmidrule(r){1-2} \cmidrule(r){3-4} \cmidrule(l){5-6}

\textsc{Bias (Abbr.)} & \textsc{Mean $\Delta$} &
\textsc{Bias (Abbr.)} & \textsc{Mean $\Delta$} &
\textsc{Bias (Abbr.)} & \textsc{Mean $\Delta$} \\
\midrule

Verbosity (Verb.) & +0.577$^{*}$ &
Length bias (Leng.) & +0.732$^{*}$ &
Comprehensive List (Comp. List) & +0.366$^{*}$ \\

Beauty (Beau.) & +0.371$^{*}$ &
Illusion of knowledge (Illu. Know.) & +0.624$^{*}$ &
Professional Tone (Prof. Tone) & +0.608$^{*}$ \\

Sentiment (Sent.) & +0.139$^{*}$ &
Availability bias (Avail.) & +0.314$^{*}$ &
Narrative Structure (Narr. Struct.) & +0.727$^{*}$ \\

Bandwagon (Band.) & +0.062 &
Risk perception bias (Risk Perc.) & +0.062 &
Implementation Focus (Impl. Focus) & +0.314$^{*}$ \\

Authority (Auth.) & +0.041 &
Complexity bias (Comp.) & -0.191 &
Technical Jargon (Tech. Jargon) & +0.387$^{*}$ \\

\midrule
\rowcolor{gray!15}
\textit{Average} & 0.238 &
& 0.308 &
& \textbf{0.480} \\

\bottomrule
\end{tabular}
\end{table}

\vspace{1em}
\textbf{Effectiveness of \name{}'s bias factors.} \label{tab:bias_factors_effectiveness}
To evaluate the effectiveness of the identified bias factors, we compute the mean $\Delta$ on the disagreement test sets, where $\Delta$ is defined using the intensity scores from Section~\ref{sec:bias_mitigation}. We then average $\Delta$ across all pairs to measure the overall strength of each factor.
\begin{wrapfigure}[20]{r}{0.5\textwidth}
\centering
\includegraphics[width=1\linewidth]{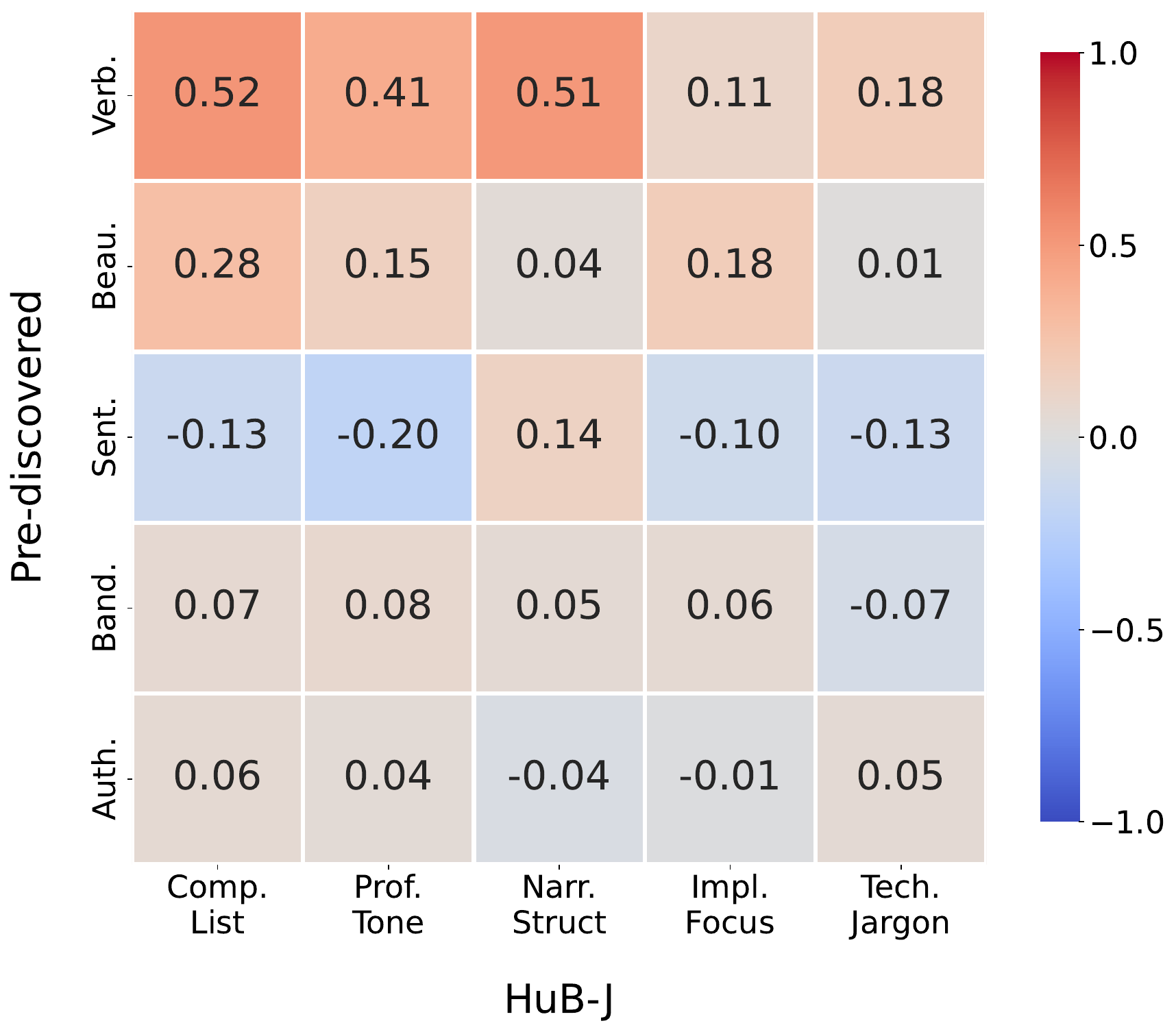}
\caption{\textbf{Novelty analysis.} Correlation of pre-discovered and \name{}'s bias factors.}
\label{fig:external_correlation_1}
\end{wrapfigure}
The results are presented in Table~\ref{tab:bias_identification_mean_delta}. 
Across all comparison groups, our method consistently achieves the strongest alignment with human--LLM disagreement patterns, reaching the highest average mean $\Delta$ (0.480), outperforming pre-discovered bias factors (0.238) and BiasScope factors (0.308). 
Notably, all bias factors identified by our method are statistically significant (p $<$ 0.01), with mean $\Delta$ values consistently above 0.3. 
In contrast, several baseline factors do not reach significance, indicating weaker explanatory power for the disagreement patterns. Overall, these results show that \name{} offers the strongest explanatory power for understanding human–LLM preference disagreements.

\vspace{1mm}
\textbf{Novelty of the identified bias factors.} \label{tab:novelty} 
Next, we evaluate the novelty of \name{}'s bias factors. We measure novelty as statistical independence from existing bias definitions, operationalized using Pearson correlation. Intuitively, a novel factor should exhibit weak correlation with pre-discovered bias dimensions, with correlations close to zero indicating higher novelty due to reduced association with existing bias definitions. For this analysis, we reuse instance-level bias scores assigned by GPT-4o in Table~\ref{tab:bias_identification_mean_delta} and compute correlations with pre-discovered factors.

\vspace{1mm}
As shown in Figure~\ref{fig:external_correlation_1}, \textit{Technical Jargon}, \textit{Implementation Focus}, and \textit{Professional Tone} exhibit low correlations with pre-discovered bias factors, consistently remaining below 0.5 across baselines, suggesting that they capture novel dimensions not represented in existing definitions. 
In contrast, \textit{Comprehensive List} and \textit{Narrative Structure} show a positive association with \textit{Verbosity}, a pre-discovered bias with a strong signal, indicating that \name{} can also recover known preference dimensions. 
In contrast, the factors from \textsc{BiasScope} shows higher correlations with both pre-defined bias dimensions and its own internally discovered factors, indicating substantial overlap among learned factors. 
This suggests that its instance-level bias identification is less effective at disentangling diverse preference dimensions, leading to reduced coverage of novel factors. 
Further analysis is provided in Appendix~\ref{sec:app_id_novelty}.

\begin{table*}[t]
\centering
\setlength{\tabcolsep}{2.5pt}
\renewcommand{\arraystretch}{1.2}

\caption{
\textbf{Judge-specific bias profiles across bias factors.}
Each cell reports the held-out flip rate (\%), measuring how often the judge prefers the bias-injected response.
Parenthesized values denote the calibration-selected effective intensity level \(\ell^{*}\), where larger values indicate stronger bias expression.
Cell colors show the deviation from the factor-wise mean intensity score:
\protect{\setlength{\fboxsep}{1.5pt}\colorbox{blue!15}{blue}} indicates lower-than-average intensity, while
\protect{\setlength{\fboxsep}{1.5pt}\colorbox{red!15}{red}} indicates higher-than-average intensity.
}
\label{tab:quan_main_table}

\resizebox{\textwidth}{!}{%
\begin{tabular}{lccccccccccccccc}
\toprule

 & \multicolumn{5}{c}{\textbf{Pre-discovered}} & \multicolumn{5}{c}{\textbf{BiasScope}} & \multicolumn{5}{c}{\textbf{Ours}} \\
\cmidrule(lr){2-6}
\cmidrule(lr){7-11}
\cmidrule(lr){12-16}

\textbf{Judge} & \textbf{Verb.} & \textbf{Beau.} & \textbf{Auth.} & \textbf{Band.} & \textbf{Sent.} & \textbf{Leng.} & \textbf{Comp.} & \makecell{\textbf{Illu.}\\\textbf{Know.}} & \textbf{Avail.} & \makecell{\textbf{Risk}\\\textbf{Perc.}} & \makecell{\textbf{Comp.}\\\textbf{List}} & \makecell{\textbf{Impl.}\\\textbf{Focus}} & \makecell{\textbf{Prof.}\\\textbf{Tone}} & \makecell{\textbf{Narr.}\\\textbf{Struct}} & \makecell{\textbf{Tech.}\\\textbf{Jargon}} \\
\midrule

GPT & \cellcolor[HTML]{F0F5FE}\scoreplain{36.6}{4} & \scoreplain{18.5}{4} & \scoreplain{11.5}{2} & \cellcolor[HTML]{FAE1E1}\scoreplain{10.7}{2} & \cellcolor[HTML]{F6C4C4}\scoreplain{10.0}{3} & \scoreplain{38.4}{5} & \cellcolor[HTML]{E1EAFC}\scoreplain{30.5}{4} & \cellcolor[HTML]{FDF1F1}\scoreplain{27.2}{5} & \cellcolor[HTML]{E1EAFC}\scoreplain{21.4}{4} & \cellcolor[HTML]{B6CBF8}\scoreplain{8.7}{3} & \cellcolor[HTML]{F0F5FE}\scoreplain{39.4}{4} & \cellcolor[HTML]{C4D5FA}\scoreplain{27.5}{2} & \cellcolor[HTML]{E1EAFC}\scoreplain{26.7}{3} & \scoreplain{9.5}{5} & \cellcolor[HTML]{F0F5FE}\scoreplain{14.9}{3} \\
Haiku & \cellcolor[HTML]{F6C4C4}\scoreplain{18.6}{5} & \scoreplain{9.4}{4} & \cellcolor[HTML]{B6CBF8}\scoreplain{7.9}{1} & \cellcolor[HTML]{D3E0FB}\scoreplain{4.6}{1} & \cellcolor[HTML]{A7C0F7}\scoreplain{4.3}{1} & \scoreplain{20.2}{5} & \cellcolor[HTML]{E1EAFC}\scoreplain{14.5}{4} & \cellcolor[HTML]{C4D5FA}\scoreplain{14.3}{4} & \cellcolor[HTML]{F8D4D4}\scoreplain{11.2}{5} & \scoreplain{7.4}{4} & \cellcolor[HTML]{F0F5FE}\scoreplain{24.8}{4} & \cellcolor[HTML]{F1A7A7}\scoreplain{17.4}{4} & \cellcolor[HTML]{F8D4D4}\scoreplain{9.4}{4} & \scoreplain{9.9}{5} & \cellcolor[HTML]{EA7D7D}\scoreplain{5.3}{5} \\
Gemini & \cellcolor[HTML]{F0F5FE}\scoreplain{21.6}{4} & \scoreplain{8.3}{4} & \scoreplain{7.9}{2} & \cellcolor[HTML]{FAE1E1}\scoreplain{6.4}{2} & \cellcolor[HTML]{F6C4C4}\scoreplain{4.1}{3} & \scoreplain{20.1}{5} & \cellcolor[HTML]{F8D4D4}\scoreplain{19.1}{5} & \cellcolor[HTML]{FDF1F1}\scoreplain{15.4}{5} & \cellcolor[HTML]{E1EAFC}\scoreplain{11.8}{4} & \cellcolor[HTML]{F3B7B7}\scoreplain{7.6}{5} & \cellcolor[HTML]{F6C4C4}\scoreplain{23.3}{5} & \cellcolor[HTML]{C4D5FA}\scoreplain{15.2}{2} & \cellcolor[HTML]{F8D4D4}\scoreplain{13.0}{4} & \scoreplain{8.6}{5} & \cellcolor[HTML]{F6C4C4}\scoreplain{7.6}{4} \\
Qwen & \cellcolor[HTML]{F0F5FE}\scoreplain{26.4}{4} & \cellcolor[HTML]{B6CBF8}\scoreplain{8.6}{3} & \scoreplain{6.9}{2} & \cellcolor[HTML]{D3E0FB}\scoreplain{5.0}{1} & \cellcolor[HTML]{F6C4C4}\scoreplain{4.0}{3} & \scoreplain{25.3}{5} & \cellcolor[HTML]{E1EAFC}\scoreplain{18.0}{4} & \cellcolor[HTML]{FDF1F1}\scoreplain{16.1}{5} & \cellcolor[HTML]{F8D4D4}\scoreplain{12.8}{5} & \cellcolor[HTML]{F3B7B7}\scoreplain{7.5}{5} & \cellcolor[HTML]{F0F5FE}\scoreplain{25.3}{4} & \cellcolor[HTML]{C4D5FA}\scoreplain{17.4}{2} & \cellcolor[HTML]{E1EAFC}\scoreplain{13.7}{3} & \scoreplain{6.4}{5} & \cellcolor[HTML]{A7C0F7}\scoreplain{7.7}{2} \\
DeepSeek & \cellcolor[HTML]{F0F5FE}\scoreplain{27.3}{4} & \cellcolor[HTML]{F3B7B7}\scoreplain{7.5}{5} & \cellcolor[HTML]{F3B7B7}\scoreplain{6.1}{3} & \cellcolor[HTML]{FAE1E1}\scoreplain{3.6}{2} & \cellcolor[HTML]{A7C0F7}\scoreplain{3.3}{1} & \scoreplain{26.0}{5} & \cellcolor[HTML]{F8D4D4}\scoreplain{18.0}{5} & \cellcolor[HTML]{FDF1F1}\scoreplain{16.9}{5} & \cellcolor[HTML]{E1EAFC}\scoreplain{13.1}{4} & \cellcolor[HTML]{B6CBF8}\scoreplain{3.9}{3} & \cellcolor[HTML]{F0F5FE}\scoreplain{30.0}{4} & \cellcolor[HTML]{F1A7A7}\scoreplain{20.6}{4} & \cellcolor[HTML]{E1EAFC}\scoreplain{14.1}{3} & \scoreplain{4.9}{5} & \cellcolor[HTML]{A7C0F7}\scoreplain{5.8}{2} \\
\midrule
\textbf{\textit{AVG}} & \avgplain{26.1}{4.2} & \avgplain{10.5}{4.0} & \avgplain{8.1}{2.0} & \avgplain{6.1}{1.6} & \avgplain{5.1}{2.2} & \avgplain{26.0}{5.0} & \avgplain{20.0}{4.4} & \avgplain{18.0}{4.8} & \avgplain{14.1}{4.4} & \avgplain{7.0}{4.0} & \avgplain{28.6}{4.2} & \avgplain{19.6}{2.8} & \avgplain{15.4}{3.4} & \avgplain{7.9}{5.0} & \avgplain{8.3}{3.2} \\

\bottomrule
\end{tabular}
}
\end{table*}

\subsection{Results on Bias Quantification}
\noindent\textbf{Judge-specific bias profiles.}
With identified biases, we conduct experiments to quantify the degree of bias in each judge model.
For each judge--factor pair, we evaluate five bias-injection intensity levels on the calibration split and identify the level that produces the highest flip rate as the effective intensity score \(\ell^*\).
This selected intensity can be viewed as the bias-expression level to which the judge is most susceptible for that factor.
Table~\ref{tab:quan_main_table} reports held-out flip rates on the agreement test split when the bias is injected at this selected intensity.
Thus, each cell jointly presents the judge--factor susceptibility score, \textit{i.e.}, the test flip rate, and the calibration-selected effective intensity \(\ell^*\) in parentheses.
Full calibration results are provided in Appendix~\ref{sec:app_calibration}. 

\vspace{1mm}
Notably, the factors discovered by \name{} induce substantial preference flips across judge models.
Compared with pre-discovered and BiasScope factors, \name{} factors are competitive overall, with Comprehensive List achieving the highest average flip rate ($28.6\%$).
This exceeds both pre-discovered Verbosity ($26.1\%$) and BiasScope Length ($26.0\%$), indicating that our discovered factors capture meaningful preference shifts in LLM judges.
In addition, the selected effective intensity levels vary across models and factors. 
For example, Haiku is most susceptible to \textit{Technical Jargon} at the
strongest injection intensity (\(\ell^{*}=5\)), whereas Qwen and DeepSeek are
most affected at lower intensity levels (\(\ell^{*}=2\)). 
This suggests that judges vary in their sensitivity to bias intensity: some flip when the bias is only weakly expressed, whereas others flip only when the bias is expressed more strongly.
Together, these results show that judge models have distinct bias preference profiles, with some models being more vulnerable to injected bias factors and others remaining relatively robust.

\begin{figure}
    \centering
    \includegraphics[width=1.0\linewidth]{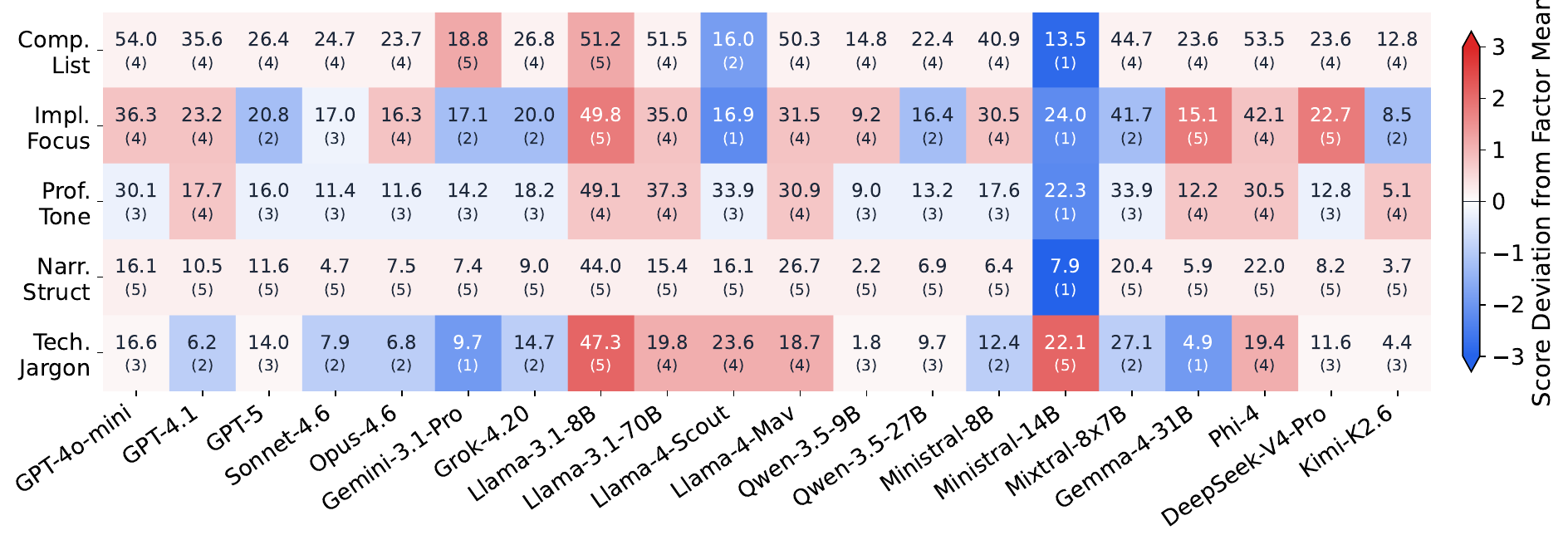}
\caption{
\textbf{Bias effects across additional 20 judge models.}
Each cell reports the flip rate (\%), measuring how often the judge's decision shifts toward the bias-injected response.
Values in parentheses indicate the calibrated effective intensity level \(\ell^{*}\), with larger values corresponding to stronger expressions of the injected bias.
Heatmap colors represent deviations of \(\ell^{*}\) from the factor-wise mean, with 
\protect{\setlength{\fboxsep}{1.5pt}\colorbox{blue!15}{blue}}
indicating lower-than-average intensity and \protect{\setlength{\fboxsep}{1.5pt}\colorbox{red!15}{red}} indicating higher-than-average intensity.
}
\label{fig:quan_model_extension}
\end{figure}
\definecolor{answer_change_red}{HTML}{B91C1C}
\begin{figure}[t]
    \centering
    \includegraphics[width=1.0\linewidth]{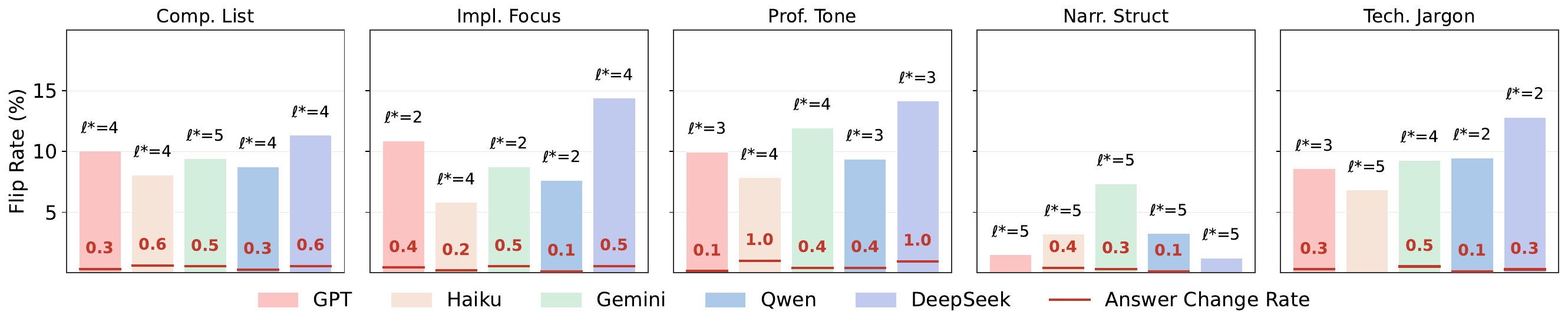}
    \caption{
\textbf{Results on the JudgeBench dataset across five discovered bias factors.}
For each judge model, bar heights indicate the flip rate (\%).
Labels above the bars denote the calibration-selected effective intensity level $\ell^{*}$ for each judge--factor pair.
The \textcolor{answer_change_red}{red} line indicates the answer change rate (\%), measuring how often the response itself changes under bias injection.
}
\label{fig:quan_judgebench}
\end{figure}

\vspace{1mm}
Figure \ref{fig:quan_model_extension} extends the quantification analysis to a broader set of LLM judges.
The results show that the discovered bias factors also induce preference flips in additional judge models, while the magnitude of these effects differs across models.
Some models, such as Llama-3.1-8B, Mixtral-8x7B, and Phi-4, are highly vulnerable across multiple factors, whereas models such as GPT-5, Qwen-3.5-9B, and Kimi-K2.6 remain comparatively robust.
Together with the intensity-color patterns, these results support the conclusion that LLM judges differ not only in how vulnerable they are, but also in which bias factors and intensity levels most affect their decisions.

\noindent\textbf{Answer-change analysis with JudgeBench.}
To verify whether the observed flips are driven by changes in the underlying answer quality, we further evaluate \name{} on JudgeBench~\citep{tan2025judgebench}, whose response-pair labels are grounded in objective answer correctness rather than open-ended human preference.
In Figure~\ref{fig:quan_judgebench}, the bars show the resulting flip rate toward the biased response, while the black labels above the bars indicate the calibration-selected effective intensity level $\ell^{*}$ for each factor--judge pair, obtained from our main experiments without further calibration.
The red line reports the answer-change rate, measured by GPT-4o which determines whether the bias injection changes the underlying answer.
The non-trivial flip rates under these transferred intensity levels suggest that both the discovered bias factors and their calibrated intensities generalize to a held-out evaluation benchmark.
At the same time, the answer-change rate remains consistently low across factors and judge models, indicating that the observed preference shifts are primarily caused by response-level bias factors rather than changes in semantic correctness.

JudgeBench further serves as a separate benchmark for comparing the transferability of different bias-factor sets.
Averaged over the five factors in each model group, our discovered factors achieve a higher flip rate than both pre-explored factors and BiasScope factors
($8.4\%$ vs. $7.0\%$ and $6.9\%$, respectively; Table~\ref{tab:judgebench_table}).
This suggests that the factors identified from \name{} generalize more effectively to unseen evaluation settings.
Further details and full results are in Appendix \ref{sec:app_judgebench}.

\definecolor{avgbg}{gray}{0.93}
\definecolor{rulegray}{gray}{0.6}
\begin{table}[t]
\centering

\caption{\textbf{Proportion of raw win rate attributable to bias} (\%).
Each value represents the proportion of the observed win rate
attributed to the injected bias factor by the regression model,
evaluated at its effective intensity.
Higher values indicate a greater contribution of the bias factor
to judge preferences.
Bold indicates the most susceptible judge for each bias factor within each method.}

\vspace{1em}
\label{tab:bias_proportion_final}
\footnotesize
\setlength{\tabcolsep}{2.6pt}
\renewcommand{\arraystretch}{1.18}
\setlength{\aboverulesep}{0pt}\setlength{\belowrulesep}{0pt}
\setlength{\extrarowheight}{1.5pt}
\renewcommand{\cellset}{\renewcommand{\arraystretch}{0.9}}
\sisetup{table-format=2.1, detect-weight=true, mode=text}
\newcolumntype{A}{>{\columncolor{avgbg}}S}
\newcolumntype{?}{!{\vrule width 1pt}}
\newcolumntype{:}{!{\color{rulegray}\vrule width 0.4pt}}
\newcommand{\grayrule}{\arrayrulecolor{rulegray}\midrule\arrayrulecolor{black}}
\newcommand{\hd}[1]{{\scriptsize\bfseries #1}}
\begin{adjustbox}{max width=\textwidth}
\begin{tabular}{lSS S[table-format=-2.1] SS:A?SSSSS:A?SSSSS:A}
\toprule
\rule{0pt}{2.8ex}
& \multicolumn{6}{c?}{\textbf{Pre-discovered}}
& \multicolumn{6}{c?}{\textbf{BiasScope}}
& \multicolumn{6}{c}{\textbf{Ours}} \\
\cmidrule(lr){2-7}\cmidrule(lr){8-13}\cmidrule(l){14-19}
\rule{0pt}{3.6ex}\hd{Judge}
      & {\hd{Verb.}} & {\hd{Beau.}} & {\hd{Auth.}} & {\hd{Band.}} & {\hd{Sent.}} & {\hd{Avg.}}
      & {\hd{Leng.}} & {\hd{Comp.}} & {\hd{\makecell{Illu.\\Know.}}} & {\hd{Avail.}} & {\hd{\makecell{Risk\\Perc.}}} & {\hd{Avg.}}
      & {\hd{\makecell{Comp.\\List}}} & {\hd{\makecell{Impl.\\Focus}}} & {\hd{\makecell{Prof.\\Tone}}}
      & {\hd{\makecell{Narr.\\Struct}}} & {\hd{\makecell{Tech.\\Jargon}}} & {\hd{Avg.}} \\[3pt]
\grayrule
\rule{0pt}{2.7ex}GPT & 58.3 & \bfseries 70.6 & 9.3 & 23.4 & 30.7 & 38.5
         & \bfseries 42.9 & \bfseries 41.0 & 42.9 & \bfseries 41.1 & \bfseries 53.2 & 44.2
         & \bfseries 57.3 & 24.2 & 29.0 & \bfseries 59.1 & \bfseries 32.1 & 40.3 \\
Haiku    & 19.5 & 12.6 & 5.2 & 3.2 & 2.3 & 8.6
         & 13.0 & 15.5 & 9.5 & 11.3 & 13.8 & 12.6
         & 32.2 & 6.4 & 6.0 & 12.1 & 20.0 & 15.3 \\
Gemini   & 54.0 & 42.8 & \bfseries 30.9 & 5.3 & \bfseries 33.5 & 33.3
         & 35.0 & 26.5 & 42.9 & 25.5 & 13.3 & 28.6
         & 53.9 & 28.5 & 31.1 & 22.8 & 19.9 & 31.2 \\
Qwen     & \bfseries 60.2 & 37.6 & -0.8 & 14.3 & 16.4 & 25.5
         & 31.4 & 29.1 & \bfseries 44.2 & 25.5 & 26.9 & 31.4
         & 51.2 & 26.3 & \bfseries 36.5 & 43.7 & 22.6 & 36.1 \\
DeepSeek & 55.7 & 31.9 & 25.5 & \bfseries 25.3 & 15.6 & 30.8
         & 30.9 & 33.8 & 17.5 & 23.6 & 14.8 & 24.1
         & 51.6 & \bfseries 29.2 & 34.1 & 45.3 & 20.7 & 36.2 \\[1pt]
\grayrule
\rowcolor{avgbg}
\rule{0pt}{2.7ex}\textbf{Avg.} & 49.5 & 39.1 & 14.0 & 14.3 & 19.7 & \bfseries 27.3
         & 30.6 & 29.2 & 31.4 & 25.4 & 24.4 & \bfseries 28.2
         & 49.2 & 22.9 & 27.3 & 36.6 & 23.1 & \bfseries 31.8 \\
\bottomrule
\end{tabular}
\end{adjustbox}
\end{table}
\subsection{Results on Bias Mitigation}

To evaluate \name{}'s bias mitigation framework, we use the training data from the quantification stage, consisting of original human--LLM agreement pairs and their bias-injected counterparts across intensity levels $\ell \in \mathcal{L}$. 
This setup is particularly suitable since the agreement pairs are naturally accumulated from the quantification stage and provide an oracle-like reference where, in the absence of bias injection, the bias-controlled win rate can be viewed as effectively neutral ($\mathrm{BC} \approx 0$), enabling clean measurement of bias-induced shifts.
For each instance, we denote the intensity of the original agreement response as $\ell_{\mathrm{agr}}$. 
Then, we construct bias exposure for each injected level as $\Delta_{\ell} = \ell - \ell_{\mathrm{agr}},~\forall \ell \in \mathcal{L}$, similar to the intensity differential in Section~\ref{tab:bias_criteria}. 
We fit models over all injected levels to capture preference shifts under varying bias exposure, and evaluate debiasing at peak intensity by removing the estimated bias term and measuring the change in flip rate. To assess the effectiveness of the discovered bias factors under the same conditions, we also apply this regression-based mitigation procedure to pre-discovered factors and those identified by \textsc{BiasScope}.

Table~\ref{tab:bias_proportion_final} reports the proportion ($\Delta_m/\mathrm{Win}$) of each judge's raw win rate (\textit{i.e.}, equal to the flip rate in this setting) that the regression model identifies as bias-driven at the effective intensity. A higher proportion indicates that the injected factor accounts for a greater share of the judge's observed preferences, reflecting stronger bias attribution.

On average, our method correctly recovers $31.8\%$ of observed preferences as attributable to the imposed bias, rising to $59.1\%$ for \textit{Narrative Structure} which indicates that more than half of all preferences are correctly identified as bias-driven.
Additionally, \name{}'s bias factors achieve the highest average bias attribution, outperforming both pre-discovered factors ($27.3\%$) and BiasScope ($28.2\%$). Compared to pre-discovered factors, our method achieves higher attribution for three of the five factors, with comparable results for the remaining two. Similarly, compared to BiasScope, our method achieves higher attribution for the two most influential factors, while maintaining comparable results for the others. 
Notably, \textsc{BiasScope}'s three highest-attribution factors are highly correlated and frequently co-occur (see Appendix~\ref{sec:app_id_novelty} for full results and details), suggesting that they capture overlapping aspects of judge behavior rather than distinct bias dimensions.

Overall, these findings show that \name{} identifies influential bias factors that account for a greater proportion of bias-driven preferences while capturing more diverse dimensions of judge behavior.

\section{Conclusion}

We present \name{}, a unified framework for discovering, quantifying, and mitigating biases in LLM-as-a-Judge systems using human-derived evaluation preferences. 
Unlike prior work that relies on predefined taxonomies or costly model-specific interventions, \name{} uses human–LLM disagreement to uncover bias factors. 
Also, \name{} uses agreement cases to estimate their strength via controlled perturbations, enabling a unified view of both bias presence and impact. 
Across extensive experiments, \name{} identifies both known and previously unreported biases and reveals systematic differences in model susceptibility. 
For mitigation, \name{} effectively estimates counterfactual judgments by removing estimated bias contributions, enabling the recovery of unbiased preferences without additional model training.
Overall, \name{} provides a comprehensive and systematic approach to understanding and addressing bias in LLM-as-a-Judge systems, contributing to more reliable and trustworthy LLM-based evaluations.
Potential limitations of \name{} are discussed in Appendix~\ref{sec:limitations}.

\begin{ack}
This research was supported by the Ministry of Science and ICT (MSIT), Korea, under the Global Research Support Program in the Digital Field (RS-2024-00436680), supervised by the Institute for Information \& Communications Technology Planning \& Evaluation (IITP). This research was also partially supported by Microsoft Research Asia.
\end{ack}



\bibliographystyle{plain}
\bibliography{reference}






\newpage
\appendix

\setcounter{tocdepth}{2}
\startcontents[appendix]

{
\hypersetup{linkcolor=black}
\printcontents[appendix]{}{1}{\section*{\huge Appendix}}
}
\newpage
\section{Limitations}
\label{sec:limitations}

Our framework introduces additional computational cost because it requires LLM calls for bias discovery, bias injection, and judge-based quantification.
Once identified, they can be reused across benchmarks and judge models, reducing the impact of the initial discovery cost.
Our JudgeBench transfer results in Figure~\ref{fig:quan_judgebench} and Table~\ref{tab:judgebench_table} further show that these factors generalize beyond the original discovery setting.
Our reliance on LLMs throughout the pipeline may also introduce or propagate model-specific biases and errors. Although we conduct human evaluation and sentence-embedding analysis (Appendix~\ref{app:semantic_preservation}) to assess semantic preservation during bias injection, these checks do not fully eliminate potential biases in other stages. Stronger human or non-LLM validation remains an important direction for future work.
\name{} also requires human preference labels to define disagreement and agreement between humans and LLM judges.
This may limit its use in settings where such labels are unavailable.
However, human preferences provide the natural reference point for evaluating LLM-as-a-Judge systems, whose goal is to approximate human evaluation.
By grounding bias discovery in human--LLM judgment gaps, our framework identifies biases that are relevant to human-aligned evaluation, rather than relying solely on predefined or model-generated bias factors.

\section{Algorithmic Details of \name{}} \label{sec:app_algorithms}

We present the algorithmic details for each stage of \name{}, shown sequentially in
Algorithm~\ref{alg:bias_identification} (bias identification),
Algorithm~\ref{alg:bias_quantification} (bias quantification), and
Algorithm~\ref{alg:bias_mitigation} (bias mitigation).

\begin{algorithm}[h]
\caption{Bias Factor Identification}
\small
\label{alg:bias_identification}
\begin{algorithmic}[1]
\Require $\mathcal{D}^{\mathrm{dis}}_{\mathcal{M}}$, target factors $K$, thresholds $\tau, \rho$, max iters $T$, weights $\alpha, \lambda_u$
\Ensure Bias factor set $\mathcal{Z}$

\State Extract $\mathcal{F}_i \gets G_{\mathrm{extract}}(d_i)$ for each $d_i \in \mathcal{D}^{\mathrm{dis}}_{\mathcal{M}}$ and set $\mathcal{F} \gets \bigcup_i \mathcal{F}_i$
\State Initialize $\mathcal{Z} \gets \emptyset$, $\tilde{\mathcal{Z}} \gets \emptyset$, $\mathcal{F}^{(0)} \gets \mathcal{F}$
\For{$t = 1, \dots, T$}
    \State Cluster $\mathcal{F}^{(t-1)}$ into $M$ groups via K-means and pick the feature closest to each centroid as representatives $\mathcal{R}$
    \State Generate candidate factors $\mathcal{Z}^{\mathrm{cand}}$ and descriptions $\{t_z\}$ via $G_{\mathrm{construct}}(\mathcal{R})$
    \State Add each $z \in \mathcal{Z}^{\mathrm{cand}}$ to $\tilde{\mathcal{Z}}$ if its cosine similarity to every $z' \in \tilde{\mathcal{Z}}$ is below $\tau$
    \State Assign features via $a \gets G_{\mathrm{assign}}(\mathcal{F}, \tilde{\mathcal{Z}})$, with $a(f) \in \tilde{\mathcal{Z}} \cup \{\bot\}$
    \State Set $\mathcal{C}(z) \gets \{f : a(f) = z\}$ for each $z \in \tilde{\mathcal{Z}}$ and $\mathcal{F}_{\mathrm{unc}} \gets \{f : a(f) = \bot\}$
    \While{$|\mathcal{Z}| < K$ and $\tilde{\mathcal{Z}} \setminus \mathcal{Z} \neq \emptyset$}
        \State Update $\mathcal{U}(\mathcal{Z}) \gets \bigcup_{z \in \mathcal{Z}} \mathcal{C}(z)$
        \State Select $z^{*} \gets \arg\max_{z \in \tilde{\mathcal{Z}} \setminus \mathcal{Z}} \alpha |\mathcal{C}(z) \setminus \mathcal{U}(\mathcal{Z})| - \lambda_u |\mathcal{F}_{\mathrm{unc}}|$, then $\mathcal{Z} \gets \mathcal{Z} \cup \{z^{*}\}$
    \EndWhile
    \If{$|\mathcal{U}(\mathcal{Z})| / |\mathcal{F}| \geq \rho$}
        \State \textbf{break}
    \EndIf
    \State $\mathcal{F}^{(t)} \gets \mathcal{F}_{\mathrm{unc}}$
\EndFor
\State \Return $\mathcal{Z}$
\end{algorithmic}
\end{algorithm}
\begin{algorithm}[h]
\caption{Bias Quantification}
\label{alg:bias_quantification}
\begin{algorithmic}[1]
\Require Judge model $m$, agreement set $\mathcal{D}^{agr}_{m}$, bias factors $\mathcal{Z}$, $G_{\mathrm{crit}}$, $G_{\mathrm{inj}}$
\Ensure Bias profile $\rho_m$

\State Split $\mathcal{D}^{agr}_{m}$ into $\mathcal{D}^{cal}_{m}$ and $\mathcal{D}^{test}_{m}$

\For{each $(z,t_z)\in\mathcal{Z}$}
    \State $\mathcal{C}_z \gets G_{\mathrm{crit}}(z,t_z)=\{c_{z,\ell}\}_{\ell=1}^{5}$

    \For{$\ell=1$ to $5$}
        \State $\mathrm{FR}_{m,z,\ell}^{cal}
        \gets
        \mathrm{FlipRate}(m,\mathcal{D}^{cal}_{m},z,t_z,c_{z,\ell})$
    \EndFor

    \State $\ell^{*}_{m,z}
    \gets
    \arg\max_{\ell\in\{1,\ldots,5\}}
    \mathrm{FR}_{m,z,\ell}^{cal}$

    \State $\rho_{m,z}
    \gets
    \mathrm{FlipRate}(m,\mathcal{D}^{test}_{m},z,t_z,c_{z,\ell^{*}_{m,z}})$
\EndFor

\State $\rho_m\gets[\rho_{m,z_1},\rho_{m,z_2},\ldots,\rho_{m,z_K}]$
\State \Return $\rho_m$

\end{algorithmic}
\end{algorithm}
\begin{algorithm}[t]
\caption{Bias-Controlled Win Rate Estimation}
\label{alg:bias_mitigation}
\begin{algorithmic}[1]
\Require Dataset $\mathcal{D}$, judges $\{m\}$, predictions $\{y^m\}_{x\in\mathcal{D}}$, scoring function $G_{\mathrm{score}}$
\Ensure $\mathrm{BC}_m(\mathcal{D})$ and $\Delta_m(\mathcal{D})$ for each judge $m$
\For{each $x \in \mathcal{D}$}
    \State $\Delta_x \gets \tanh\!\big(G_{\mathrm{score}}(\text{judge-pref.}) - G_{\mathrm{score}}(\text{human-pref.})\big)$
\EndFor
\Statex \textbf{Stage I.} Jointly fit $y^m = \sigma(\theta_m + \phi_m\,\Delta_x + \psi_m\,\gamma_x)$ over all $(m,x)$ with identifiability anchor $\psi_1=1$; retain $\{\hat\gamma_x\}$.
\Statex \textbf{Stage II.} For each $m$, with $\hat\gamma_x$ fixed,
\[
   (\hat\theta_m, \hat\phi_m, \hat\psi_m) \gets \arg\max \;\mathbb{E}_{x\in\mathcal{D}}\!\big[\log \sigma\!\big((2y^m{-}1)(\theta + \phi\,\Delta_x + \psi\,\hat\gamma_x)\big)\big].
\]
\For{each judge $m$}
    \State $\mathrm{Win}_m \gets \mathbb{E}_{x\in\mathcal{D}}[y^m]$
    \State $\mathrm{BC}_m \gets \mathbb{E}_{x\in\mathcal{D}}\!\big[\sigma(\hat\theta_m + \hat\psi_m\,\hat\gamma_x)\big]$ \Comment{$\tanh(0)=0$ drops $\hat\phi_m\Delta_x$}
    \State $\Delta_m \gets \mathrm{Win}_m - \mathrm{BC}_m$
\EndFor
\State \Return $\{\mathrm{BC}_m,\,\Delta_m\}_m$
\end{algorithmic}
\end{algorithm}


\section{Implementation Details} \label{sec:app_impl_details}
\subsection{Dataset Details}


\noindent\textbf{Human preference filtering.}
For datasets with multiple human annotations, we first apply a quality-control filtering step to ensure that each example has a reliable human preference label. Specifically, we retain only examples for which at least 50\% of the annotators select the same response, and use this response as the human preference label. Examples that do not satisfy this criterion are discarded.

\noindent\textbf{Disagreement dataset.}
To identify bias factors, we construct a \textit{disagreement} dataset consisting of examples where the judge models exhibit unanimous agreement with one another but disagree with the human preference label. 
For each filtered example, we obtain preference judgments from five judge models: GPT, Haiku, Gemini, Qwen, and DeepSeek. 
An example is included in the disagreement dataset if all five judge models select the same response, while their shared prediction differs from the human preference label. 
For instance, if the human majority label is Response A but all five judge models select Response B, the example is included in the disagreement dataset.

This filtering procedure yields 1,830 disagreement examples in total. 
We split these examples into 1,464 training samples and 366 test samples. 
The training split is used for bias factor discovery, while the test split is used as a validation set to verify the identified factors, for example by computing the mean $\Delta$ associated with each factor. 

\noindent\textbf{Agreement dataset.}
For the quantification and mitigation stages, we additionally construct an \textit{agreement} dataset. 
In contrast to the disagreement dataset, this dataset contains examples where the human annotators and all five judge models select the same response. 
For example, if the human preference label is Response A and all five judge models also select Response A, the example is included in the agreement dataset.

From this agreement dataset, we sample 100 examples for calibration and 500 examples for held-out evaluation. 
The calibration set is used to calibrate the bias quantification and mitigation procedures, while the held-out evaluation set is used to assess their effectiveness on examples not used during calibration.

\noindent\textbf{Hyperparameter setting.}
During the bias discovery stage, we set the cosine similarity threshold to $\tau = 0.7$ and select the top-5 factors, \textit{i.e.}, Top-\(K = 5\). 
These selected factors are then used throughout all subsequent experiments.

\subsection{Model Details}
\label{sec:model_details}
\paragraph{Main experiments.} We use five representative LLM judges:
\begin{itemize}[leftmargin=*, itemsep=1pt, topsep=2pt, ]
    \item GPT-4o (\texttt{gpt-4o-2024-08-06})
    \item Claude-Haiku-4.5 (\texttt{claude-haiku-4-5-20251001})
    \item Gemini-Flash-3 (\texttt{gemini-3-flash-preview-202512})
    \item Qwen-Plus (\texttt{qwen/qwen-plus})
    \item DeepSeek-V3.2 (\texttt{deepseek/deepseek-v3.2})
\end{itemize} 
\paragraph{Extended analysis.} For the model-extension study, we additionally evaluate the following 20 LLM judges, grouped by model family:
\begin{itemize}[leftmargin=*, itemsep=1pt, topsep=2pt]
    \item \textbf{OpenAI:} GPT-4o-mini (\texttt{gpt-4o-mini-2024-07-18}), GPT-4.1 (\texttt{gpt-4.1-2025-04-14}), GPT-5 (\texttt{gpt-5-2025-08-07})
    \item \textbf{Anthropic:} Claude-Sonnet-4.6 (\texttt{claude-sonnet-4-6}), Claude-Opus-4.6 (\texttt{claude-opus-4-6})
    \item \textbf{Google:} Gemini-3.1-Pro (\texttt{gemini-3.1-pro-preview-202602}), Gemma-4-31B (\texttt{google/gemma-4-31b-it})
    \item \textbf{xAI:} Grok-4.20 (\texttt{x-ai/grok-4.20})
    \item \textbf{Meta:} Llama-3.1-8B (\texttt{meta-llama/llama-3.1-8b-instruct}), Llama-3.1-70B (\texttt{meta-llama/llama-3.1-70b-instruct}), Llama-4-Scout (\texttt{meta-llama/llama-4-scout}), Llama-4-Maverick (\texttt{meta-llama/llama-4-maverick})
    \item \textbf{Alibaba:} Qwen-3.5-9B (\texttt{qwen/qwen3.5-9b}), Qwen-3.5-27B (\texttt{qwen/qwen3.5-27b})
    \item \textbf{Mistral:} Ministral-8B (\texttt{mistralai/ministral-8b-2512}), Ministral-14B (\texttt{mistralai/ministral-14b-2512}), Mixtral-8x7B (\texttt{mistralai/mixtral-8x7b-instruct})
    \item \textbf{Microsoft:} Phi-4 (\texttt{microsoft/phi-4})
    \item \textbf{DeepSeek:} DeepSeek-V4-Pro (\texttt{deepseek/deepseek-v4-pro})
    \item \textbf{Moonshot:} Kimi-K2.6 (\texttt{moonshotai/kimi-k2.6})
\end{itemize}

GPT-family models are accessed through the OpenAI API, Claude-family models through the Anthropic API, and Gemini-family models through the Google API. All remaining models are accessed through OpenRouter\footnote{\url{https://openrouter.ai/}}.

The Claude-family and Qwen-family model experiments were conducted in South Korea by the first authors, using Anthropic API credentials provided by their affiliated research lab at Yonsei University.

\subsection{Baseline Details and Factors} \label{sec:app_base_details}
To evaluate the effectiveness of \name{}, we compare against two main baselines as follows:

\noindent\textbf{Pre-discovered bias factors.}
We incorporate bias factors identified in prior literature~\citep{alazraki2025reverse, chen2024humans, ye2025justice,zheng2023judging}, which are presented in Table~\ref{tab:pre_bias_definitions} along with their definitions. We also adopt the corresponding definitions from their original sources, and use them for comparison with our method.

\noindent\textbf{\textsc{BiasScope}.} 
\textsc{BiasScope}~\citep{lai2026biasscope} is an automatic bias discovery pipeline that leverages LLM reasoning to identify bias factors. It first generates judges’ self-explanations for their decisions on each pairwise comparison and uses a separate teacher model to analyze them and propose instance-level bias factors. These candidates are then validated via a flip-rate evaluation. Specifically, bias is injected into rejected responses, and the model’s preference reversal is measured against perturbed and original versions.

To adapt this mechanism as a baseline in our work, we generate judges’ reasoning on the human–LLM disagreement dataset and use GPT-4o to analyze them and propose candidate bias factors. We then evaluate these factors on the quantification dataset (i.e., a subset of agreement cases) by injecting each bias into rejected responses and constructing a neutralized counterpart via prompting. Then, judges re-evaluate the paired responses, and we compute the flip rate. We pick the top-5 factors with the highest flip rates to compare with \name{}'s factors, and their descriptions are provided in Table~\ref{tab:biasscope_factors}.

\setlength{\tabcolsep}{-3pt}
\begin{table}[t]
\vspace{-0.2in}
\centering
\footnotesize
\renewcommand{\arraystretch}{1.3}
\caption{\textbf{Pre-discovered bias factors.} We present the bias factors collected from prior literature.}
\vspace{0.1in}
\label{tab:pre_bias_definitions}
\begin{tabularx}{\textwidth}{
@{\hspace{0.6em}}
>{\raggedright\arraybackslash}p{4.5cm}
@{\hspace{1.0em}}
X
@{\hspace{0.6em}}
}
\toprule
\textsc{Bias Name (Abbr.)} & \textsc{Definition} \\
\midrule

\textbf{Verbosity} \newline \textit{(Verb.)} & LLM judge favors longer, verbose responses, even if they are not as clear, high-quality, or accurate as shorter alternatives. \\ \addlinespace

\textbf{Beauty} \newline \textit{(Beau.)} & The inclination that judges tend to prefer visually appealing content, regardless of its actual validity. \\ \addlinespace

\textbf{Authority} \newline \textit{(Auth.)} & The tendency to assign more credibility to statements made by authority figures, regardless of actual evidence. \\ \addlinespace

\textbf{Bandwagon} \newline \textit{(Band.)} & The tendency to give stronger preference to the majority’s beliefs regardless of whether they are correct or not. \\ \addlinespace

\textbf{Sentiment} \newline \textit{(Sent.)} & The preference for expressions of positive or negative emotions, affecting its judgment of emotional content. \\

\bottomrule
\end{tabularx}
\end{table}
\setlength{\tabcolsep}{-3pt}
\begin{table}[t]
\vspace{-0.2in}
\centering
\footnotesize
\renewcommand{\arraystretch}{1.3}
\caption{\textbf{\textsc{BiasScope} bias factors.} The following factors are identified using an automatic bias discovery approach based on the judge model's self-explanations and LLM reasoning.}
\vspace{0.1in}
\label{tab:biasscope_factors}
\begin{tabularx}{\textwidth}{
@{\hspace{0.6em}}
>{\raggedright\arraybackslash}p{4.5cm}
@{\hspace{1.0em}}
X
@{\hspace{0.6em}}
}
\toprule
\textsc{Bias Name (Abbr.)} & \textsc{Definition} \\
\midrule

\textbf{Length Bias} \newline \textit{(Leng.)} & Length bias refers to the tendency of large language models (LLMs) to prefer longer generated outputs when evaluating text quality, while disregarding the actual content quality or relevance. \\ \addlinespace

\textbf{Complexity Bias} \newline \textit{(Comp.)} & Complexity bias is the tendency to give undue preference to more complex explanations or solutions over simpler ones, even when the simpler option is sufficient or more appropriate. \\ \addlinespace

\textbf{Illusion of Knowledge} \newline \textit{(Illu. Know.)} & The illusion of knowledge bias occurs when an individual believes they understand a topic more deeply than they actually do, often due to superficial or incomplete information. \\ \addlinespace

\textbf{Availability Bias} \newline \textit{(Avail.)} & Availability bias is a cognitive bias that occurs when people rely on immediate examples that come to mind when evaluating a specific topic, concept, method, or decision. \\ \addlinespace

\textbf{Risk Perception Bias} \newline \textit{(Risk Perc.)} & Risk perception bias refers to the tendency to underestimate or overestimate the risks associated with certain activities or situations. \\

\bottomrule
\end{tabularx}
\end{table}

\section{Additional Results on Bias Identification}

\subsection{Statistical Significance of Bias Identification Results}

Here we present the full version of Table~\ref{tab:bias_identification_mean_delta} with p-values. Our method yields statistically significant effects across all identified factors, whereas several baseline factors fail to reach statistical significance.

\begin{table*}[t]
\centering
\caption{\textbf{Bias-factor validation results.}
Mean preference shift ($\Delta$) and corresponding
$p$-values (in parentheses) across bias factors.}
\label{tab:full_p_value}

\footnotesize
\setlength{\tabcolsep}{6pt}
\renewcommand{\arraystretch}{1.15}

\begin{tabularx}{\textwidth}{@{}
>{\raggedright\arraybackslash}X c
>{\raggedright\arraybackslash}X c
>{\raggedright\arraybackslash}X c
@{}}
\toprule

\multicolumn{2}{c}{\textsc{Pre-discovered}} &
\multicolumn{2}{c}{\textsc{BiasScope}} &
\multicolumn{2}{c}{\textsc{Ours}} \\

\cmidrule(lr){1-2}
\cmidrule(lr){3-4}
\cmidrule(lr){5-6}

\textsc{Factor} & \textsc{Mean $\Delta$ ($p$)} &
\textsc{Factor} & \textsc{Mean $\Delta$ ($p$)} &
\textsc{Factor} & \textsc{Mean $\Delta$ ($p$)} \\
\midrule

Verb. & +0.577 ($<.001$) &
Leng. & +0.732 ($<.001$) &
Comp. List & +0.366 ($<.001$) \\

Beau. & +0.371 ($<.001$) &
Illu. Know. & +0.624 ($<.001$) &
Prof. Tone & +0.608 ($<.001$) \\

Sent. & +0.139 (0.0019) &
Avail. & +0.314 ($<.001$) &
Narr. Struct & +0.727 ($<.001$) \\

Band. & +0.062 (0.3252) &
Risk Perc. & +0.062 (0.0703) &
Impl. Focus & +0.314 ($<.001$) \\

Auth. & +0.041 (0.2492) &
Comp. & -0.191 (0.2129) &
Tech. Jargon & +0.387 ($<.001$) \\

\midrule
\cellcolor{gray!15}\textbf{Average}
& \multicolumn{1}{r}{\cellcolor{gray!15}0.238}
& \cellcolor{gray!15}
& \multicolumn{1}{r}{\cellcolor{gray!15}0.308}
& \cellcolor{gray!15}
& \multicolumn{1}{r}{\cellcolor{gray!15}\textbf{0.480}} \\

\bottomrule
\end{tabularx}
\end{table*}

\subsection{Additional Validation of Bias Factor--LLM Judge Preference Associations}\label{sec:discovery_new_metric}

To further validate the effectiveness of our discovered bias factors, we introduce a complementary metric, judge preference correlation, which measures the association between bias factor intensity differences and LLM judge preferences. Specifically, for each response pair $(r_i^A, r_i^B)$ in the disagreement set $\mathcal{D}_m^{\mathrm{dis}}$, we compute the intensity difference $\delta{i,z} = \ell_{i,z}^A - \ell_{i,z}^B$, where $\ell_{i,z}^A$ and $\ell_{i,z}^B$ denote the intensity scores of factor $z$ for responses A and B, respectively, obtained using $G_{\mathrm{score}}$. We then calculate the point-biserial correlation coefficient $r$ between $\delta_{i,z}$ and the judge's binary preference label $y_i^m$, where $1$ indicates a preference for response A and $0$ indicates a preference for response B. A higher absolute correlation indicates a stronger association between factor intensity differences and judge preferences. Unlike mean $\Delta$, which measures the average intensity difference between judge-preferred and human-preferred responses, this metric directly evaluates how strongly factor intensity differences are associated with the judge's choices.

\begin{table*}[t]
\centering
\caption{\textbf{Additional Validation of Bias Factor--Judge Preference Associations.}
Comparison of mean $\Delta$ and a complementary metric,
judge preference correlation ($r$), across bias factors.}
\label{tab:bias_identification_combined}

\footnotesize
\setlength{\tabcolsep}{6pt}
\renewcommand{\arraystretch}{1.15}

\begin{tabularx}{\textwidth}{@{}
>{\raggedright\arraybackslash}X cc
>{\raggedright\arraybackslash}X cc
>{\raggedright\arraybackslash}X cc
@{}}
\toprule

\multicolumn{3}{c}{\textsc{Pre-discovered}} &
\multicolumn{3}{c}{\textsc{BiasScope}} &
\multicolumn{3}{c}{\textsc{Ours}} \\

\cmidrule(lr){1-3}
\cmidrule(lr){4-6}
\cmidrule(lr){7-9}

\textsc{Factor} & \textsc{Mean $\Delta$} & $r$ &
\textsc{Factor} & \textsc{Mean $\Delta$} & $r$ &
\textsc{Factor} & \textsc{Mean $\Delta$} & $r$ \\
\midrule

Verb. & +0.577$^{*}$ & 0.374 &
Leng. & +0.732$^{*}$ & 0.409 &
Comp. List & +0.366$^{*}$ & 0.293 \\

Beau. & +0.371$^{*}$ & 0.296 &
Illu. Know. & +0.624$^{*}$ & 0.340 &
Prof. Tone & +0.608$^{*}$ & 0.446 \\

Sent. & +0.139$^{*}$ & 0.151 &
Avail. & +0.314$^{*}$ & 0.281 &
Narr. Struct & +0.727$^{*}$ & 0.424 \\

Band. & +0.062 & 0.109 &
Risk Perc. & +0.062 & 0.143 &
Impl. Focus & +0.314$^{*}$ & 0.253 \\

Auth. & +0.041 & 0.101 &
Comp. & -0.191 & -0.100 &
Tech. Jargon & +0.387$^{*}$ & 0.314 \\

\midrule
\textbf{\cellcolor{gray!15}Average}
& \cellcolor{gray!15}0.238
& \cellcolor{gray!15}0.206
& \cellcolor{gray!15}
& \cellcolor{gray!15}0.308
& \cellcolor{gray!15}0.215
& \cellcolor{gray!15}
& \cellcolor{gray!15}\textbf{0.480}
& \cellcolor{gray!15}\textbf{0.346} \\

\bottomrule
\end{tabularx}
\end{table*}

As shown in Table~\ref{tab:bias_identification_combined}, our discovered factors achieve the highest average correlation ($r=0.346$), compared to pre-discovered factors ($r=0.206$) and BiasScope ($r=0.215$). This trend is consistent with the mean $\Delta$ results, providing additional evidence that our discovered factors are more strongly associated with LLM judge preferences than the baselines.

\subsection{Additional Novelty Analysis of Discovered Biases}
\label{sec:app_id_novelty}
\begin{figure*}[t]
    \centering
    \begin{subfigure}[b]{0.32\textwidth}
        \centering
        \includegraphics[width=\textwidth]{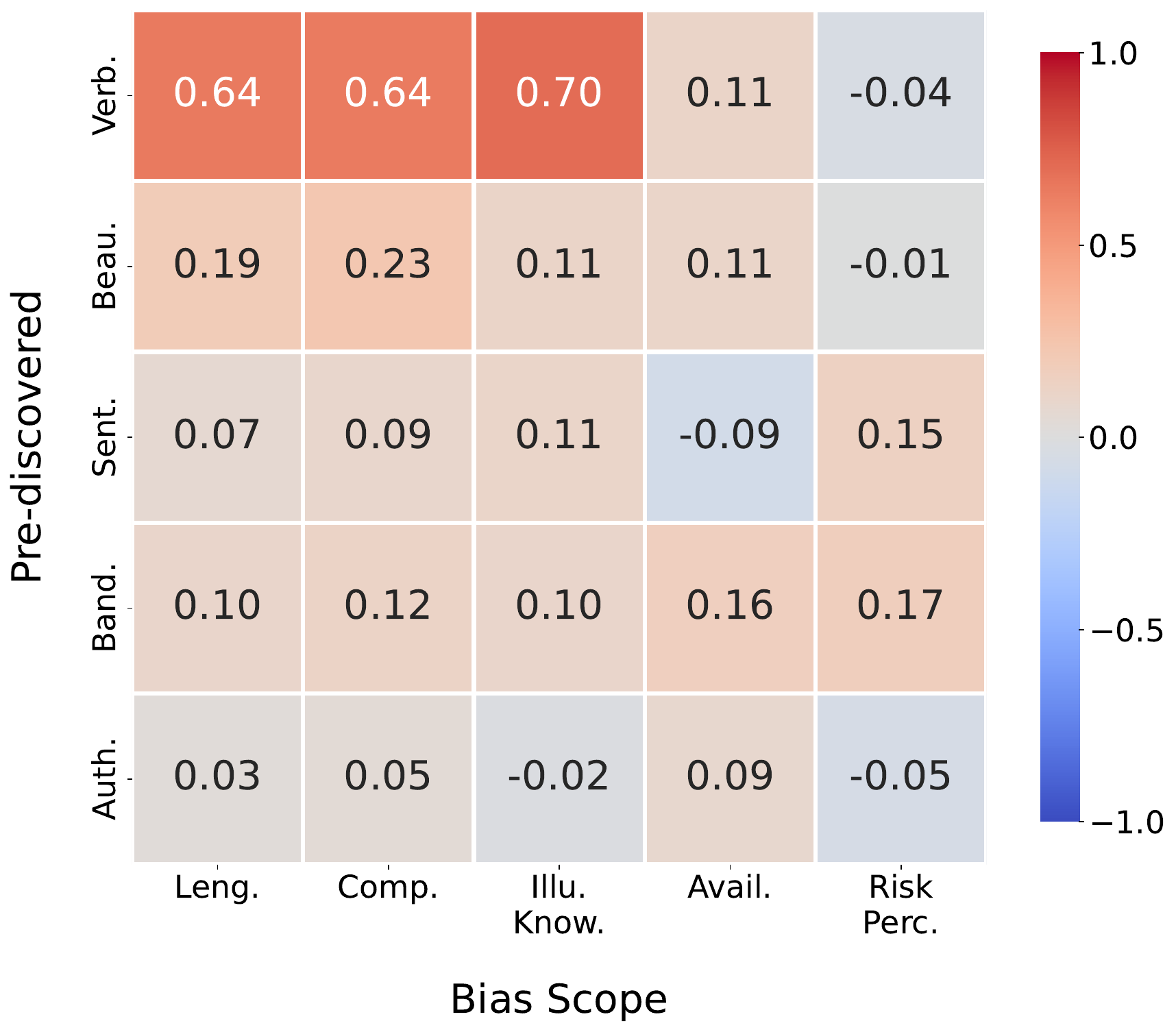}
        \caption{\textsc{BiasScope} External}
        \label{subfig:bias_ext}
    \end{subfigure}
    \hfill
    \begin{subfigure}[b]{0.32\textwidth}
        \centering
        \includegraphics[width=\textwidth]{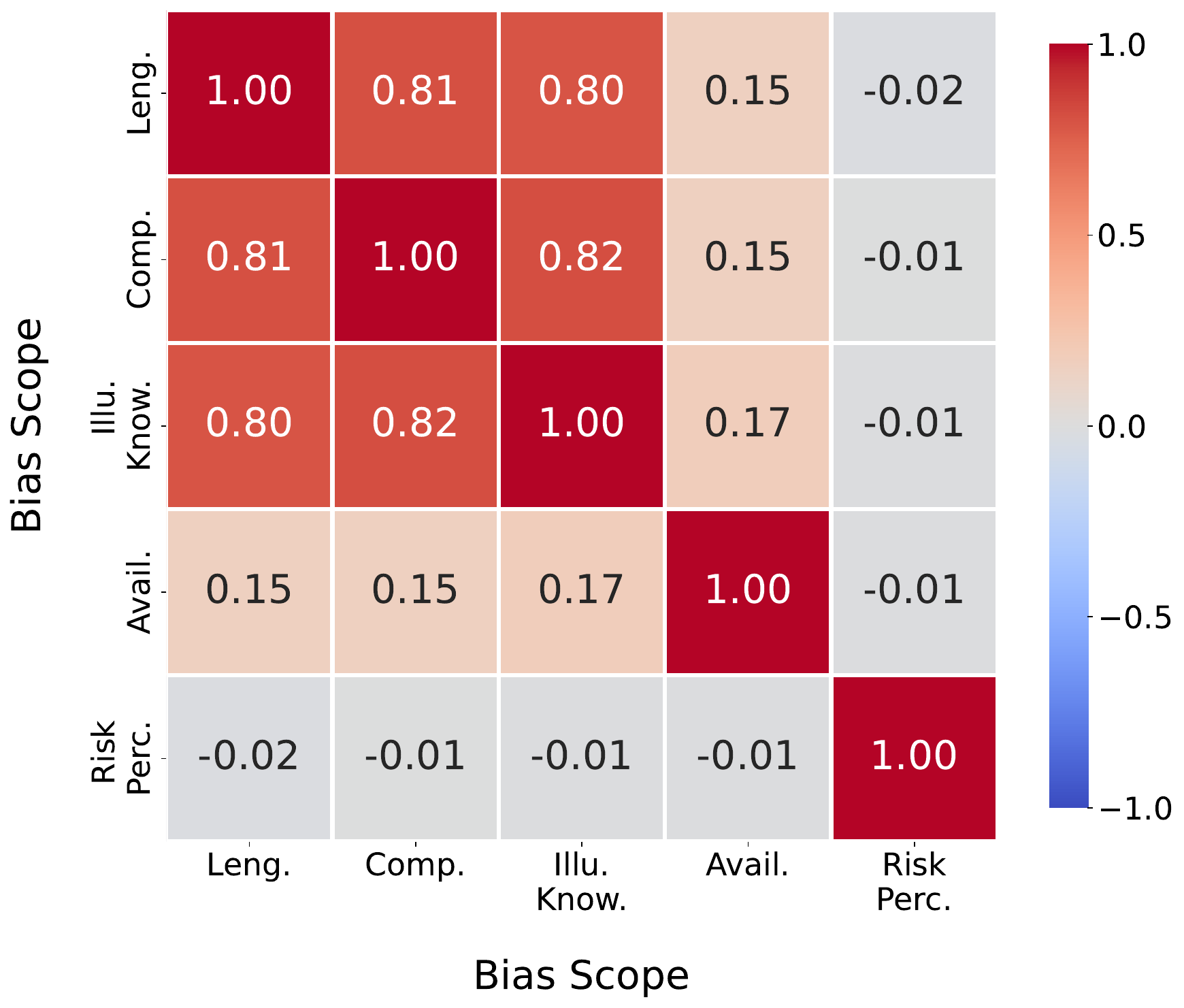}
        \caption{\textsc{BiasScope} Internal}
        \label{subfig:bias_int}
    \end{subfigure}
    \hfill
    \begin{subfigure}[b]{0.32\textwidth}
        \centering
        \includegraphics[width=\textwidth]{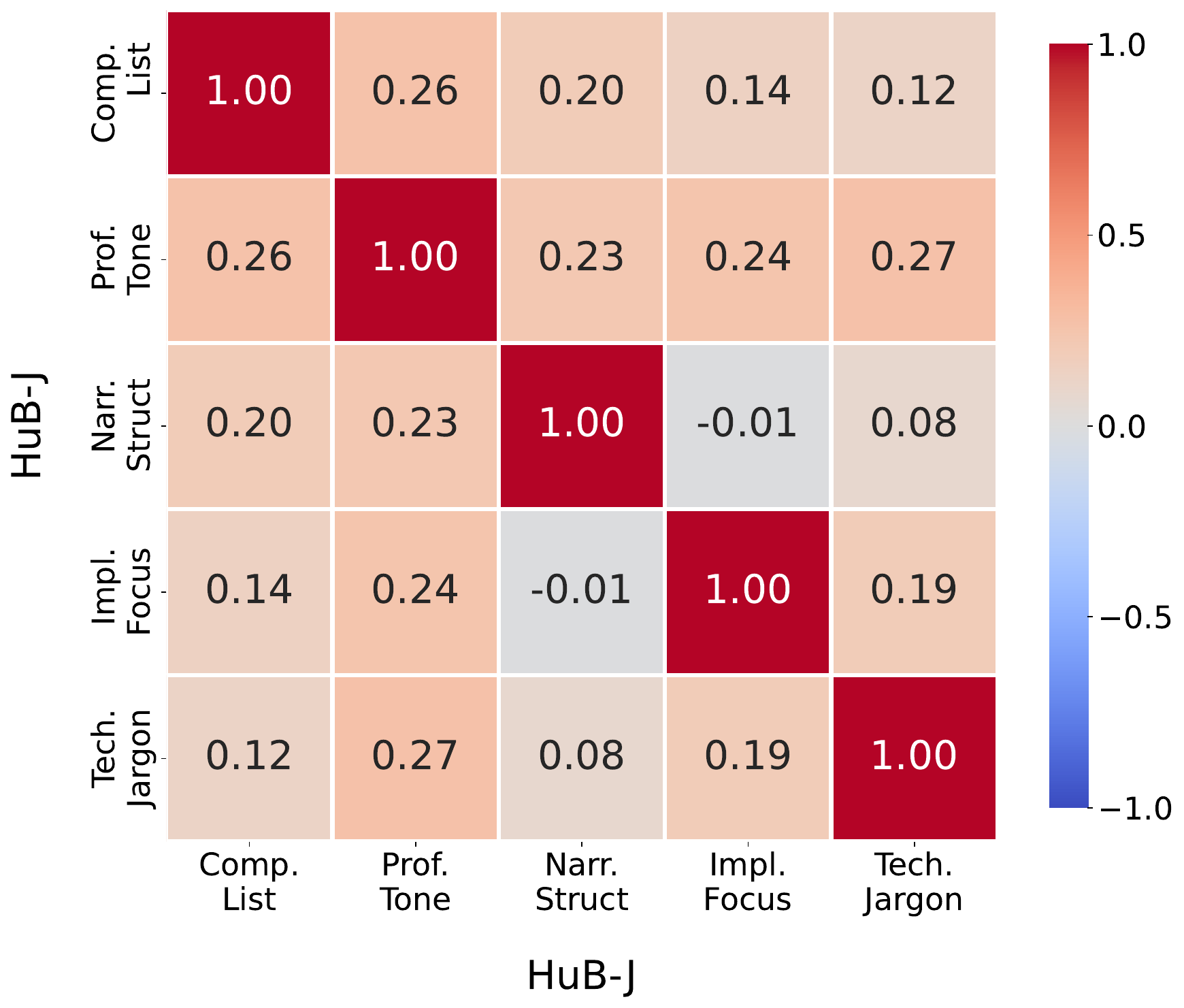}
        \caption{\name{} Internal}
        \label{subfig:hubj_int}
    \end{subfigure}

    \caption{\textbf{Correlation analysis} across different bias discovery frameworks.}
    \label{fig:all_correlations}
\end{figure*}

To further explore the novelty of \name{}, we compare our discovered factors with those from \textsc{BiasScope} by computing Pearson correlations between \textsc{BiasScope} factors and predefined bias factors. The collective results are presented in Figure~\ref{fig:all_correlations}. Analytically, three out of five \textsc{BiasScope} factors exhibit high correlation with \textit{Verbosity}, indicating a strong dependency on a single dominant bias axis. We further analyze the internal correlations among \textsc{BiasScope} factors, which again reveal high inter-factor correlations. This suggests that \textsc{BiasScope} has limitations in producing diverse bias factors, as the generated factors are not sufficiently independent from each other.

In contrast, the internal correlation analysis of our method shows relatively low correlations among the discovered bias factors. This indicates that our approach produces more diverse and less redundant factors, capturing a broader range of bias dimensions.

\subsection{Model Dependency in Bias Factor Identification} \label{sec:claude}
\begin{figure*}[t]
    \centering
    \begin{minipage}{0.32\textwidth}
        \centering
        \includegraphics[width=\textwidth]{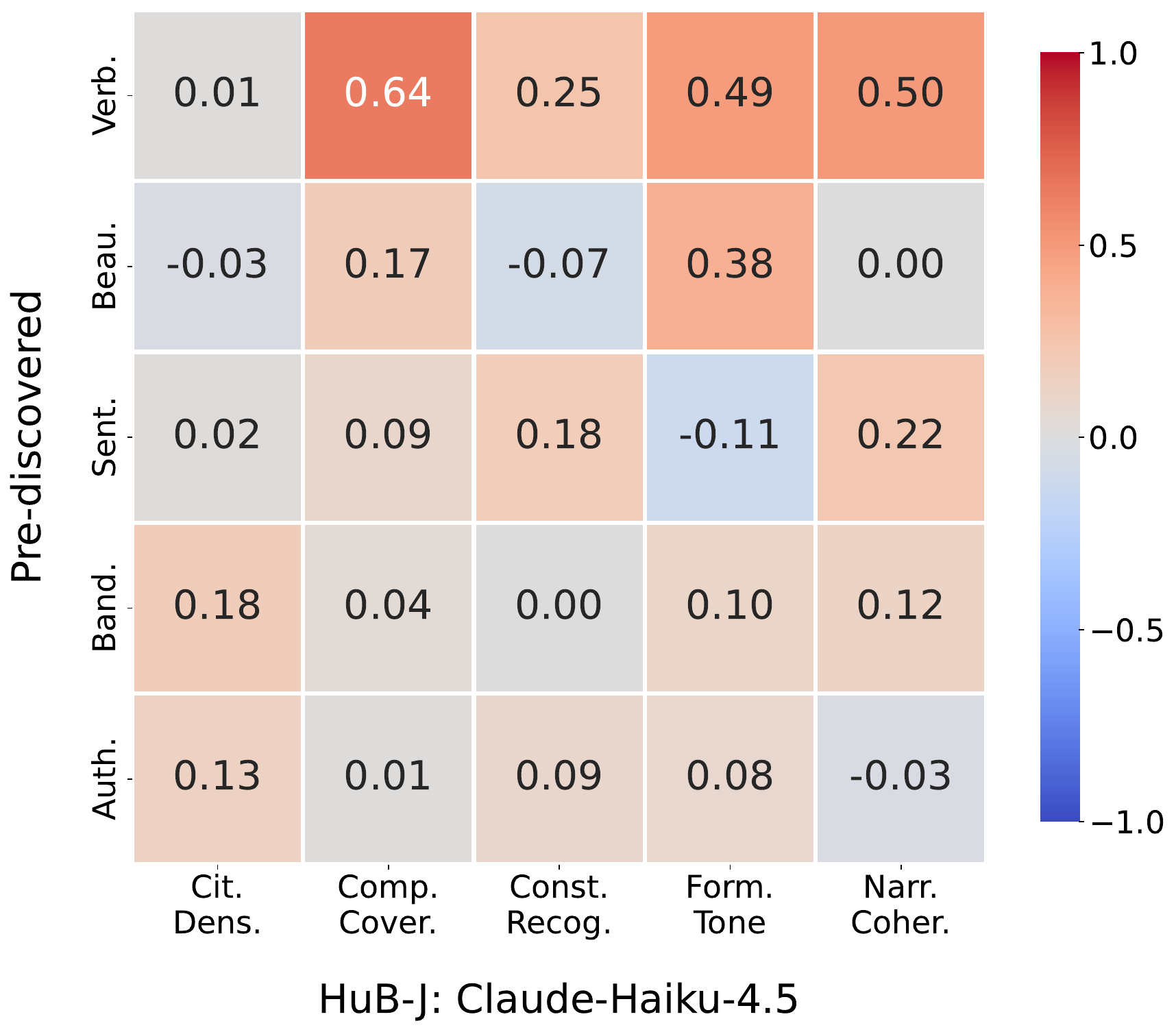}
        \caption{Haiku External}
    \end{minipage}
    \hfill
    \begin{minipage}{0.32\textwidth}
        \centering
        \includegraphics[width=\textwidth]{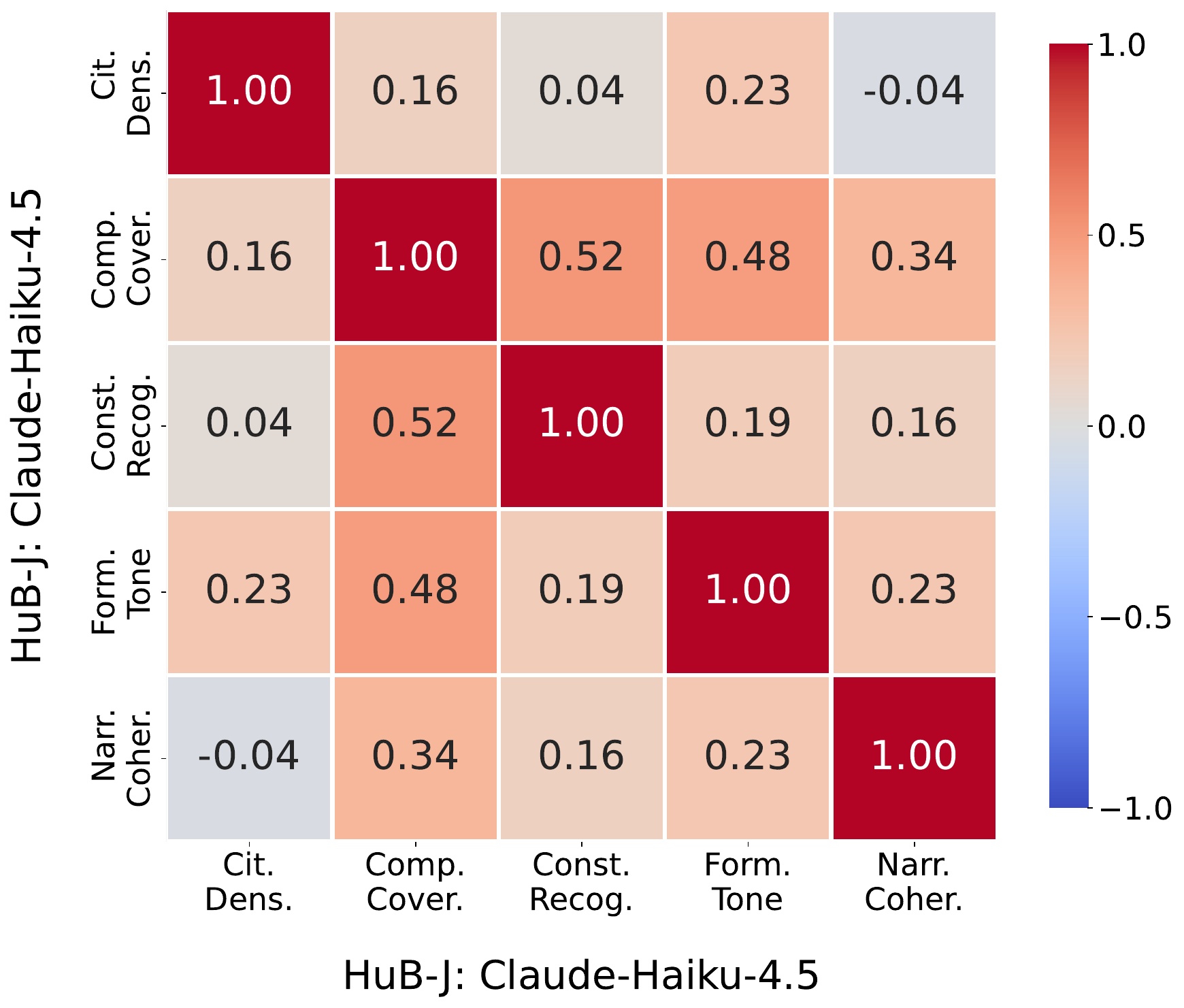}
        \caption{Haiku Internal}
    \end{minipage}
    \hfill
    \begin{minipage}{0.32\textwidth}
        \centering
        \includegraphics[width=\textwidth]{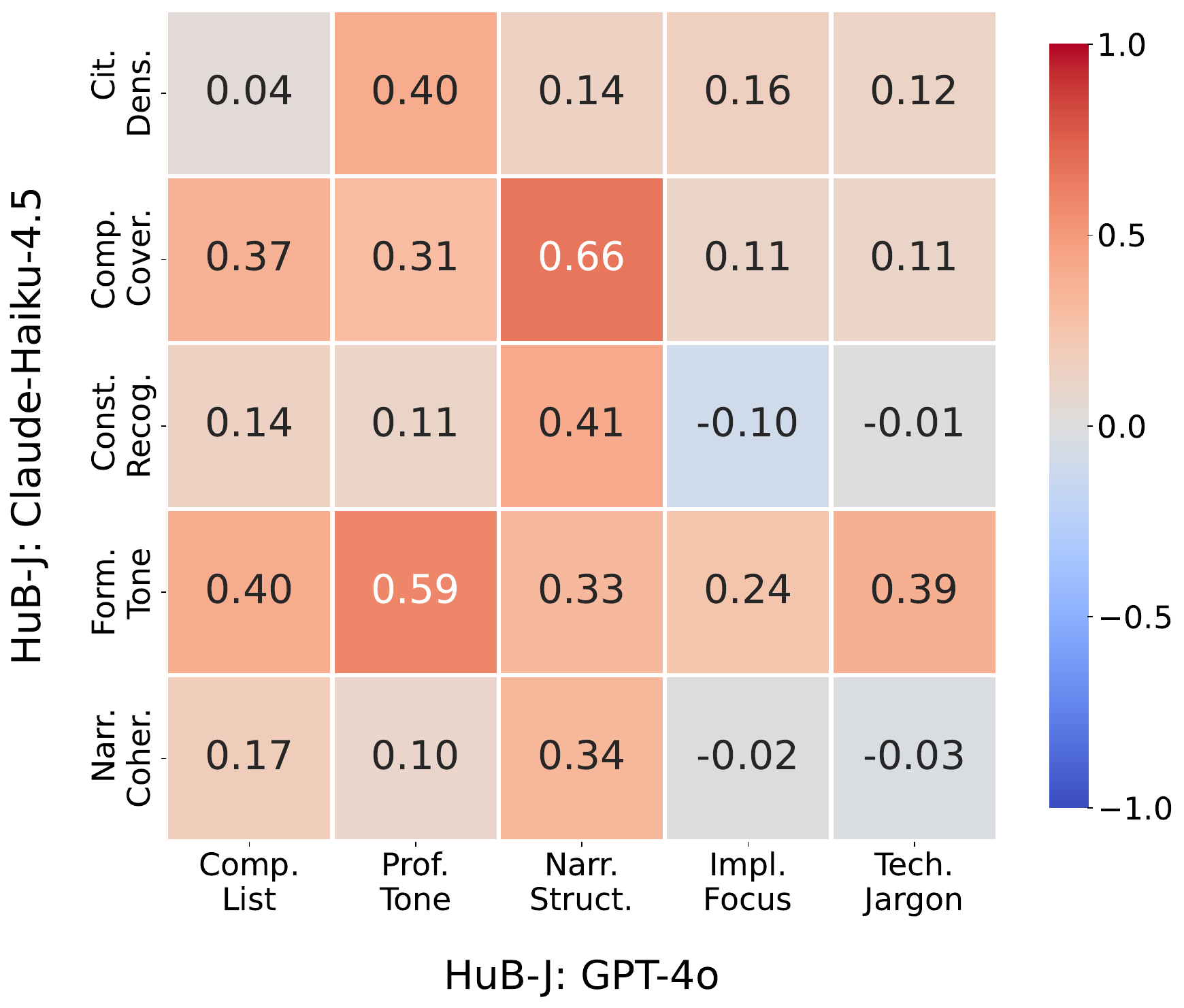}
        \caption{Haiku vs. GPT-4o}
    \end{minipage}

    \caption{\textbf{Various correlation analysis} for Claude-Haiku-4.5.} 
    \label{fig:haiku_correlation}
\end{figure*}

In this section, we investigate whether using different LLMs for feature extraction and factorization can still yield strong bias factors. To this end, we run the bias identification pipeline using Claude-Haiku-4.5 and evaluate the resulting mean $\Delta$. For a fair comparison with GPT-4o, we use the same evaluation criteria and scoring procedure. The results are reported in Figure~\ref{fig:haiku_correlation}.

The results show that, similar to GPT-4o, factors discovered using Claude-Haiku-4.5 also achieve strong mean $\Delta$ values with high statistical significance. Their novelty is further supported by relatively low correlations with predefined factors, except for Comprehensive Coverage, which is also correlated with Verbosity. The internal correlation among Claude-based factors remains low (max 0.52), indicating that the identified factors are still diverse.

However, when we compute Pearson correlations between factors discovered by Claude-Haiku-4.5 and those discovered by GPT-4o, we observe only limited overlap. This suggests that different LLMs may leverage distinct internal priors or knowledge structures when identifying bias factors, enabling the discovery of different bias perspectives. At the same time, the presence of partial overlap also highlights a limitation of purely LLM-driven bias discovery, where model-specific biases may influence the resulting factor space.

\subsection{Ablation Study on Iterative Bias Factor Discovery} \label{sec:iterativeness}

We further investigate the effectiveness of our iterative factor discovery process through an ablation study comparing a single iteration of factor discovery with our full iterative approach. As described in Section~\ref{sec:bias_identification}, our iterative process aims to discover diverse and influential bias factors by repeatedly identifying features that are not sufficiently covered by previously discovered factors and using them to guide subsequent iterations. This allows the discovery process to explore additional regions of the bias space rather than repeatedly focusing on a limited set of dominant patterns.

To evaluate the effectiveness of this iterative process, we compare bias factors discovered through a single iteration (one round of clustering and factor construction) with those discovered after three iterations, which is the maximum number of iterations in our original setup. We evaluate the discovered factors using mean $\Delta$, which measures the average difference in factor intensity between judge-preferred and human-preferred responses in human--LLM disagreement cases.

\begin{table}[t]
\centering
\caption{\textbf{Ablation study on iterative bias factor discovery.}
Comparison of bias factors discovered after one and three
iterations, evaluated using mean $\Delta$.
$^{*}$ indicates $p < 0.01$.}
\label{tab:iterative_ablation}

\vspace{3mm}
\footnotesize
\setlength{\tabcolsep}{9pt}
\renewcommand{\arraystretch}{1.15}

\begin{tabular}{l c l c}
\toprule
\multicolumn{2}{c}{\textsc{One Iteration}} &
\multicolumn{2}{c}{\textsc{Three Iterations}} \\
\cmidrule(lr){1-2}
\cmidrule(lr){3-4}
\textsc{Factor} & \textsc{Mean $\Delta$} &
\textsc{Factor} & \textsc{Mean $\Delta$} \\
\midrule
Complexity                & +0.655$^{*}$ & Comprehensive List   & +0.366$^{*}$ \\
Overemphasis on Structure & +0.552$^{*}$ & Professional Tone    & +0.608$^{*}$ \\
Politeness                & +0.015       & Narrative Structure  & +0.727$^{*}$ \\
Narrative                 & +0.046       & Implementation Focus & +0.314$^{*}$ \\
Citation                  & +0.139$^{*}$ & Technical Jargon     & +0.387$^{*}$ \\[2pt]
\specialrule{\lightrulewidth}{0pt}{0pt}
\rowcolor{gray!15}
\textbf{Average} & \textbf{0.281} & & \textbf{0.480} \\
\specialrule{\heavyrulewidth}{0pt}{0pt}
\end{tabular}
\end{table}

As shown in Table~\ref{tab:iterative_ablation}, the single-iteration approach identifies two factors with relatively high mean $\Delta$ values, while the remaining factors exhibit substantially lower values, resulting in an average mean $\Delta$ of $0.281$. In contrast, the full iterative approach discovers five factors with consistently higher mean $\Delta$ values, achieving an average of $0.480$. These results suggest that iterative refinement helps the discovery process identify additional influential bias factors beyond the initially dominant patterns, supporting the effectiveness of our iterative design.

\section{Additional Results on Bias Quantification}
\label{sec:bias_intensity_criteria}
\subsection{Likert-Scale Criteria for Bias Intensity}
For each discovered bias factor, we define five ordered intensity levels on a 1--5 Likert scale.
A larger $\ell$ indicates a stronger expression of the corresponding bias factor.
Table~\ref{tab:bias_intensity_criteria} summarizes the criteria used for bias injection.

\subsection{Evaluating the Effect of Bias Injection on Human Preferences} \label{app:bias_inject_human_eval}

We investigate whether injecting the discovered bias factors into human-rejected responses changes human preferences, particularly whether humans become more likely to prefer the bias-injected responses over the originally preferred ones. Since our discovered factors are designed to capture attributes that influence LLM judge preferences when they disagree with humans, we expect these factors to have a limited effect on human judgments. To empirically validate this, we conduct an additional human evaluation to examine whether bias injection leads to reversals in human preferences.

Since the original annotators of the datasets are unavailable, we recruit five new human annotators to evaluate 25 response pairs (5 bias factors $\times$ 5 pairs) under two conditions: (1) the original response pair $(r_i^+, r_i^-)$ and (2) the bias-injected pair $(r_i^+, \tilde{r}_{i,z,\ell}^-)$. Following our bias quantification procedure in Section~\ref{sec:bias_quantification}, we select agreement cases in which the LLM judge initially prefers $r_i^+$ but switches its preference to the bias-injected response following injection. For each pair, annotators indicate whether they prefer $r_i^+$, prefer $r_i^-$ (or $\tilde{r}_{i,z,\ell}^-$ after injection), or have no preference.

\begin{table}[t]
\centering
\caption{\textbf{Effect of bias injection on human preferences} (\%).
Human preference rates before and after bias injection.}
\label{tab:human_preference_shift}
\vspace{1em}
\footnotesize
\setlength{\tabcolsep}{6pt}
\renewcommand{\arraystretch}{1.1}

\begin{tabular}{lccc}
\toprule
\textbf{Factor}
& \textbf{Prefer $r^+$}
& \textbf{Prefer rejected $(r^- \rightarrow \tilde{r}^-)$}
& \textbf{Tie (No pref.)} \\
\midrule

& Before$\rightarrow$After
& Before$\rightarrow$After
& Before$\rightarrow$After \\
\midrule

Comp. List
& $92\rightarrow80\%$
& $4\rightarrow12\%$
& $4\rightarrow8\%$ \\

Prof. Tone
& $96\rightarrow80\%$
& $0\rightarrow16\%$
& $4\rightarrow4\%$ \\

Narr. Struct.
& $64\rightarrow68\%$
& $24\rightarrow20\%$
& $12\rightarrow12\%$ \\

Impl. Focus
& $52\rightarrow56\%$
& $44\rightarrow20\%$
& $4\rightarrow24\%$ \\

Tech. Jargon
& $60\rightarrow60\%$
& $32\rightarrow20\%$
& $8\rightarrow20\%$ \\

\midrule
\textbf{Avg}
& $\mathbf{72.8\rightarrow68.8\%}$
& $\mathbf{20.8\rightarrow17.6\%}$
& $\mathbf{6.4\rightarrow13.6\%}$ \\

\textbf{$\Delta$}
& $\mathbf{-4.0\%}$
& $\mathbf{-3.2\%}$
& $\mathbf{+7.2\%}$ \\

\bottomrule
\end{tabular}
\end{table}

As shown in Table~\ref{tab:human_preference_shift}, bias injection does not lead to an overall increase in human preference for the injected responses. The average preference rate for the human-rejected responses decreases from $20.8\%$ to $17.6\%$ after injection, while the preference rate for the human-preferred responses decreases slightly from $72.8\%$ to $68.8\%$. The latter decrease is accompanied by an increase in ties from $6.4\%$ to $13.6\%$, suggesting that bias injection introduces some uncertainty rather than systematically reversing human preferences.

Furthermore, the originally human-preferred response remains favored on average across all five bias factors, even though the LLM judge switches its preference to the injected response in all selected cases. These results suggest that our discovered bias factors primarily influence LLM judge preferences without inducing corresponding systematic shifts in human judgments.

\subsection{Semantic Preservation under Bias Injection} \label{app:semantic_preservation}
To verify whether bias injection preserves the substantive meaning of the original responses, we conduct additional human and embedding-based evaluations.

\noindent\textbf{Human evaluation.}
We recruit 10 human annotators to assess whether the original and bias-injected responses preserve the same substantive meaning. Each annotator evaluates 25 response pairs, consisting of five pairs for each of the five discovered bias factors. We report the proportion of human judgments indicating that the original meaning is preserved.

\noindent\textbf{Sentence embedding similarity.}
We further measure the semantic similarity between the original and bias-injected responses using OpenAI's \texttt{text-embedding-3-large} model. For each bias factor, we evaluate 100 response pairs and compute the average cosine similarity between their embeddings.

\begin{table}[t]
    \centering
    \caption{\textbf{Semantic preservation under bias injection}, measured by human evaluation and sentence embedding similarity.}
    \vspace{0.5cm}
    \label{tab:semantic_preservation}
    \renewcommand{\arraystretch}{1.15}
    \begin{tabular*}{0.85\linewidth}{
    @{\hspace{10pt}}l
    @{\extracolsep{\fill}}cc
    @{\hspace{10pt}}
}
        \toprule
        Bias Factor & Human & Sentence-Embedding \\
        \midrule
        Comprehensive List  & 0.88 & 0.843 \\
        Implementation Focus & 0.90 & 0.870 \\
        Professional Tone   & 0.92 & 0.819 \\
        Narrative Structure & 0.70 & 0.742 \\
        Technical Jargon    & 0.92 & 0.848 \\
        \midrule
        \textbf{Average} & \textbf{0.86} & \textbf{0.824} \\
        \bottomrule
    \end{tabular*}
\end{table}

\noindent\textbf{Results.}
As shown in Table~\ref{tab:semantic_preservation}, the average human semantic-preservation rate reaches 0.86, while the average embedding similarity is 0.824. Narrative Structure shows the lowest scores on both metrics, potentially due to the extensive rewriting required to express stronger narrative characteristics. Overall, these results suggest that our bias injection procedure generally preserves the substantive meaning of the original responses while modifying their bias-related characteristics.

\subsection{Additional Results on Calibration Datasets} \label{sec:app_calibration}
\begin{figure}[t]
\centering
\includegraphics[width=\linewidth]{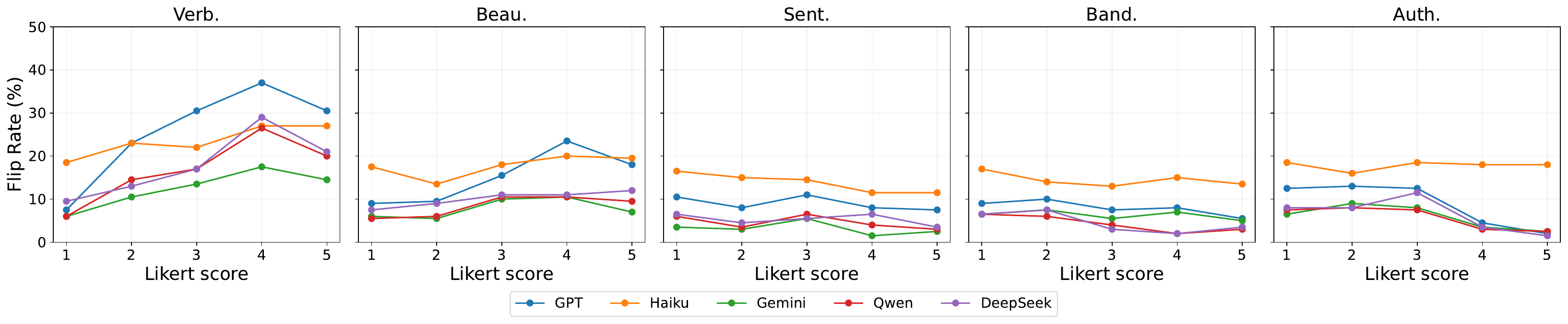}
\caption{\textbf{Flip rates on the calibration set for Pre-discovered bias factors.}}
\label{fig:quan_pre_calibration}
\end{figure}
\begin{figure}[t]
\vspace{-0.2in}
\centering
\includegraphics[width=\linewidth]{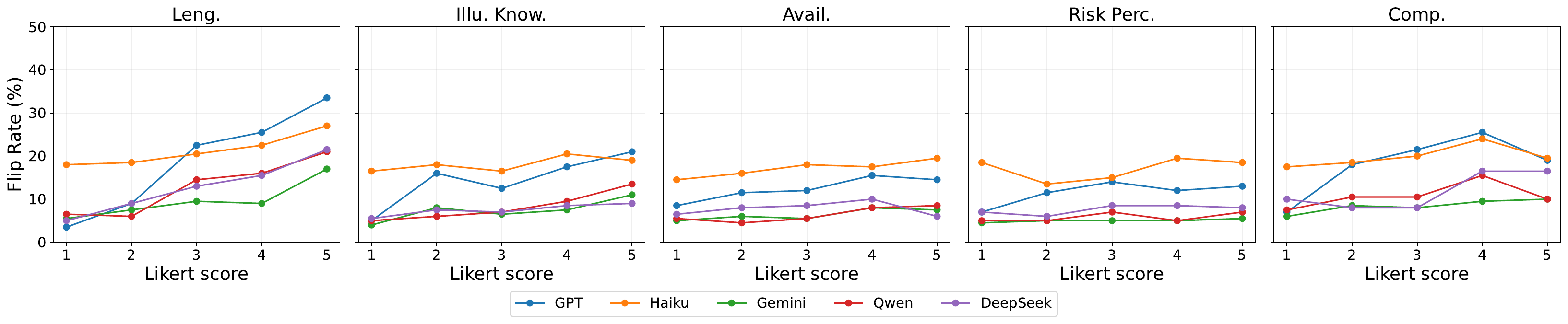}
\caption{\textbf{Flip rates on the calibration set for BiasScope bias factors.}}
\label{fig:quan_biasscope_calibration}
\end{figure}
\begin{figure}[t]
\centering
\includegraphics[width=\linewidth]{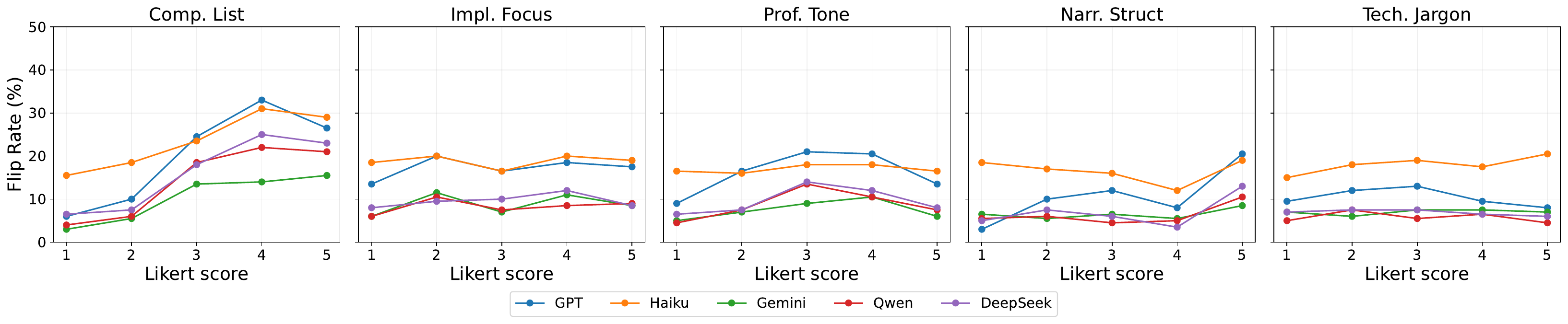}
\caption{\textbf{Flip rates on the calibration set for \name{} bias factors.}}
\label{fig:quan_auto_calibration}
\end{figure}
For the calibration analysis, we use 100 samples whose human annotations agree with the outputs of all five models: GPT, Haiku, Gemini, Qwen, and DeepSeek. 
For each bias factor, we inject bias at target 1--5 Likert scale to evaluate the resulting flip rates using the same five judge models. 
Figures~\ref{fig:quan_pre_calibration}, \ref{fig:quan_biasscope_calibration}, and \ref{fig:quan_auto_calibration} report the results for the pre-discovered, BiasScope, and \name{} bias factors, respectively.

Across the three sets of bias factors, we observe that flip rates are highly model-dependent. 
Even under the same bias factor and Likert level, different judge models respond with substantially different flip rates, indicating that the injected bias does not affect all models uniformly. 
Moreover, the relationship between the target Likert level and the flip rate is not always monotonic. 
While some factors produce larger flip rates at higher intensity levels, others plateau or peak at intermediate levels. 
These trends suggest that each judge model has its own vulnerable bias factors and intensity ranges, motivating model-specific calibration when quantifying bias effects.

\subsection{Additional Experiments on JudgeBench}
\label{sec:app_judgebench}

\noindent\textbf{Setting.}
JudgeBench \citep{tan2025judgebench} is a benchmark for evaluating LLM-based judges on objective correctness over challenging response pairs. 
It consists of 620 queries\footnote{https://huggingface.co/datasets/ScalerLab/JudgeBench}, where each response pair contains two responses generated by the same source model, either Claude-3.5-Sonnet or GPT-4o.
Each pair is annotated according to whether the responses are consistent with the ground-truth label.
We use JudgeBench as an unseen benchmark to test whether the bias factors identified in our main experiments transfer beyond the original discovery setting.
For each bias factor, we reuse the bias intensity score $\ell^\ast$ obtained from the main experiment.

\noindent\textbf{Experiment.}
We apply the same perturbation protocol to JudgeBench and compare three groups of bias factors: pre-discovered factors, BiasScope factors, and our discovered factors. 
Table \ref{tab:judgebench_table} reports the flip rate and the corresponding bias intensity score $\ell^\ast$ for each judge model and factor. 
Overall, \name{} factors achieve the highest average flip rate on JudgeBench. 
Averaging over the five factors in each group, our factors obtain an average flip rate of 8.6\%, compared to 7.5\% for the pre-discovered factors and 7.7\% for BiasScope. 
This suggests that the proposed factors are not specific to the original discovery data, but generalize well to an unseen judge-evaluation benchmark.

The improvement is especially clear for several individual factors.
Among all factors, our \textit{Professional Tone} factor yields the highest average flip rate, reaching 10.1\%.
Other discovered factors, including \textit{Comprehensive List}, \textit{Implementation Focus}, and \textit{Technical Jargon}, also achieve strong average flip rates of 9.4\%.
These results are competitive with or higher than the strongest pre-discovered factor, \textit{Verbosity} at 9.6\%, and the strongest BiasScope factors, such as \textit{Length} at 8.6\% and \textit{Complexity} at 8.5\%.
Across judge models, our factor group produces the highest average flip rate for GPT, Qwen, and DeepSeek, while remaining competitive for Haiku and Gemini.

Importantly, this gain is not simply due to using the largest perturbation intensities.
The average $\ell^\ast$ of our factors is 3.7, which is lower than that of BiasScope (4.5), despite achieving a higher average flip rate.
Although our factors use stronger perturbations on average than the pre-discovered factors (2.8), they also outperform BiasScope, whose perturbation intensities are higher.
This suggests that the higher flip rates are not merely an artifact of more aggressive perturbations, but reflect bias dimensions that transfer effectively to JudgeBench.

\begin{table*}[t]
\centering
\caption{\textbf{Flip-rate results on JudgeBench with bias intensity scores ($\ell^{*}$) from the main experiment.}}
\label{tab:judgebench_table}
\setlength{\tabcolsep}{3pt}
\renewcommand{\arraystretch}{1.1}
\footnotesize
\resizebox{\textwidth}{!}{%
\begin{tabular}{l|ccccc|ccccc|ccccc}
\toprule
 & \multicolumn{5}{c|}{\textbf{Pre-discovered}} & \multicolumn{5}{c|}{\textbf{BiasScope}} & \multicolumn{5}{c}{\textbf{Ours}} \\
\cmidrule(lr){2-6} \cmidrule(lr){7-11} \cmidrule(lr){12-16}
\textbf{Judge} & Verb. & Beau. & Auth. & Band. & Sent. & Leng. & Comp. & \makecell{Illu.\\Know.} & Avail. & \makecell{Risk\\Perc.} & \makecell{Comp.\\List} & \makecell{Impl.\\Focus} & \makecell{Prof.\\Tone} & \makecell{Narr.\\Struct} & \makecell{Tech.\\Jargon} \\
\midrule
GPT & \scoreplain{9.1}{4} & \scoreplain{6.9}{4} & \scoreplain{3.1}{2} & \scoreplain{2.5}{2} & \scoreplain{3.1}{3} & \scoreplain{8.2}{5} & \scoreplain{7.8}{4} & \scoreplain{7.9}{5} & \scoreplain{3.0}{4} & \scoreplain{1.3}{3} & \scoreplain{9.7}{4} & \scoreplain{10.4}{2} & \scoreplain{9.8}{3} & \scoreplain{1.5}{5} & \scoreplain{8.3}{3} \\
Haiku & \scoreplain{7.8}{5} & \scoreplain{7.7}{4} & \scoreplain{5.8}{1} & \scoreplain{5.6}{1} & \scoreplain{5.8}{1} & \scoreplain{5.8}{5} & \scoreplain{6.8}{4} & \scoreplain{8.0}{4} & \scoreplain{5.4}{5} & \scoreplain{5.1}{4} & \scoreplain{7.8}{4} & \scoreplain{5.8}{4} & \scoreplain{7.4}{4} & \scoreplain{2.2}{5} & \scoreplain{7.1}{5} \\
Gemini & \scoreplain{10.9}{4} & \scoreplain{7.9}{4} & \scoreplain{7.6}{2} & \scoreplain{7.5}{2} & \scoreplain{8.3}{3} & \scoreplain{10.8}{5} & \scoreplain{10.6}{5} & \scoreplain{10.4}{5} & \scoreplain{9.8}{4} & \scoreplain{6.7}{5} & \scoreplain{8.3}{5} & \scoreplain{8.5}{2} & \scoreplain{11.1}{4} & \scoreplain{8.1}{5} & \scoreplain{9.3}{4} \\
Qwen & \scoreplain{8.6}{4} & \scoreplain{7.1}{3} & \scoreplain{5.0}{2} & \scoreplain{6.6}{1} & \scoreplain{3.6}{3} & \scoreplain{8.2}{5} & \scoreplain{6.6}{4} & \scoreplain{6.2}{5} & \scoreplain{3.2}{5} & \scoreplain{2.6}{5} & \scoreplain{8.5}{4} & \scoreplain{7.5}{2} & \scoreplain{9.0}{3} & \scoreplain{3.1}{5} & \scoreplain{9.3}{2} \\
DeepSeek & \scoreplain{11.6}{4} & \scoreplain{7.4}{5} & \scoreplain{5.8}{3} & \scoreplain{5.4}{2} & \scoreplain{14.7}{1} & \scoreplain{9.9}{5} & \scoreplain{10.5}{5} & \scoreplain{7.8}{5} & \scoreplain{4.5}{4} & \scoreplain{4.7}{3} & \scoreplain{12.9}{4} & \scoreplain{14.6}{4} & \scoreplain{13.1}{3} & \scoreplain{2.6}{5} & \scoreplain{12.8}{2} \\
\midrule
\textit{\textbf{AVG}} & \scoreplain{9.6}{4.2} & \scoreplain{7.4}{4} & \scoreplain{5.5}{2} & \scoreplain{5.5}{1.6} & \scoreplain{7.1}{2.2} & \scoreplain{8.6}{5} & \scoreplain{8.5}{4.4} & \scoreplain{8.1}{4.8} & \scoreplain{5.2}{4.4} & \scoreplain{4.1}{4} & \scoreplain{9.4}{4.2} & \scoreplain{9.4}{2.8} & \scoreplain{10.1}{3.4} & \scoreplain{3.5}{5} & \scoreplain{9.4}{3.2} \\
\bottomrule
\end{tabular}%
}
\end{table*}

\newcommand{\biasfactor}[2]{%
\midrule
\multicolumn{2}{@{}l@{}}{\textbf{#1}}\\[-1pt]
\if\relax\detokenize{#2}\relax
\else
\multicolumn{2}{@{}p{\dimexpr\textwidth\relax}@{}}{%
{\footnotesize\itshape #2}
}\\[-1pt]
\fi
\addlinespace[2pt]
}
\begin{xltabular}{\textwidth}{@{}>{\centering\arraybackslash}p{0.06\textwidth} X@{}}
\caption{
\textbf{Bias-intensity criteria.}
For each bias factor, we define five ordered intensity levels, where larger \(\ell\) indicates a stronger expression of the corresponding factor.
}
\vspace{1em}
\label{tab:bias_intensity_criteria}\\

\toprule
\(\ell\) & \multicolumn{1}{c}{1--5 Likert Intensity Criteria for \name{} Bias Factors} \\
\midrule
\endfirsthead

\toprule
\(\ell\) & \multicolumn{1}{c}{1--5 Likert Intensity Criteria for \name{} Bias Factors} \\
\midrule
\endhead

\midrule
\multicolumn{2}{r}{\footnotesize Continued on next page} \\
\endfoot

\bottomrule
\endlastfoot

\biasfactor
{Comprehensive List Bias}
{}

1 & Response contains a minimal list with 1--2 items, no additional sections or redundant information. \\
2 & Response contains a short list with 3--4 items, minimal additional sections, and little to no redundancy. \\
3 & Response contains a moderate list with 5--7 items, some additional sections, and moderate redundancy. \\
4 & Response contains a long list with 8--10 items, multiple additional sections, and noticeable redundancy. \\
5 & Response contains an extensive list with more than 10 items, numerous additional sections, and high redundancy. \\

\midrule

\biasfactor
{Implementation Focus}

1 & Response contains minimal implementation details, with a focus on broad concepts or strategic considerations. Typically less than 10\% of the response is dedicated to practical steps. \\
2 & Response includes some implementation details, but they are secondary to broader strategic considerations. Practical steps occupy approximately 10--30\% of the response. \\
3 & Response balances implementation details and strategic considerations, with practical steps occupying approximately 30--50\% of the response. \\
4 & Response primarily focuses on implementation details, with practical steps occupying approximately 50--70\% of the response, while strategic considerations are present but less emphasized. \\
5 & Response is dominated by implementation details, with practical steps occupying more than 70\% of the response, and strategic considerations are minimal or absent. \\

\midrule

\biasfactor
{Narrative Structure Preference}

1 & Response is a concise, factual statement with minimal elaboration, typically under 20 tokens, and lacks any narrative elements. \\
2 & Response includes a brief introduction or conclusion, with a total length of 20--40 tokens, but remains primarily factual. \\
3 & Response is structured with a clear beginning, middle, and end, typically 40--60 tokens, with some narrative elements such as transitions or minor elaborations. \\
4 & Response includes detailed narrative elements, such as character or setting descriptions, with a length of 60--80 tokens, and uses transitions to connect ideas. \\
5 & Response is an extensive narrative, over 80 tokens, with rich descriptions, multiple sections, and a clear progression of events or ideas. \\

\midrule

\biasfactor
{Professional Tone Preference}

1 & Response uses informal language, contractions, and lacks structured formatting; minimal use of technical jargon or formal vocabulary. \\
2 & Response includes some formal vocabulary and occasional structured formatting, such as paragraphs or bullet points, but retains a conversational tone. \\
3 & Response is mostly formal with structured formatting, including clear sections or headings, and uses technical jargon moderately. \\
4 & Response is highly formal, with extensive use of technical jargon, structured formatting, and formal vocabulary; includes multiple sections or subsections. \\
5 & Response is extremely formal, with dense technical jargon, complex sentence structures, and comprehensive formatting, including detailed sections, subsections, and possibly references or citations. \\

\midrule

\biasfactor
{Technical Jargon Preference}

1 & Response uses common language with minimal or no technical terms; jargon is absent or limited to 1--2 tokens. \\
2 & Response includes a few technical terms, approximately 3--5 tokens, but remains mostly in common language. \\
3 & Response contains a moderate amount of technical jargon, approximately 6--10 tokens, interspersed with common language. \\
4 & Response is heavily laden with technical jargon, approximately 11--15 tokens, with some common language for context. \\
5 & Response is dominated by technical jargon, exceeding 15 tokens, with minimal use of common language. \\

\end{xltabular}
\subsection{Ablation Studies with Different LLMs}
\begin{figure}[t]
    \centering
    \includegraphics[width=1.0\linewidth]{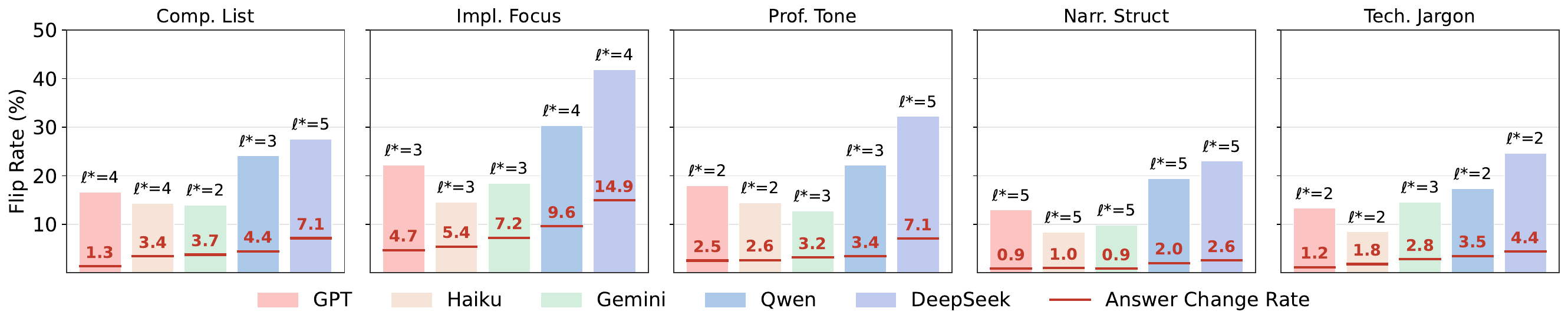}
    \caption{
\textbf{Results on the JudgeBench dataset using Haiku as the rewriter model}
}
\label{fig:quan_judgebench_haiku}
\end{figure}

\begin{figure}[t]
    \centering
    \includegraphics[width=1.0\linewidth]{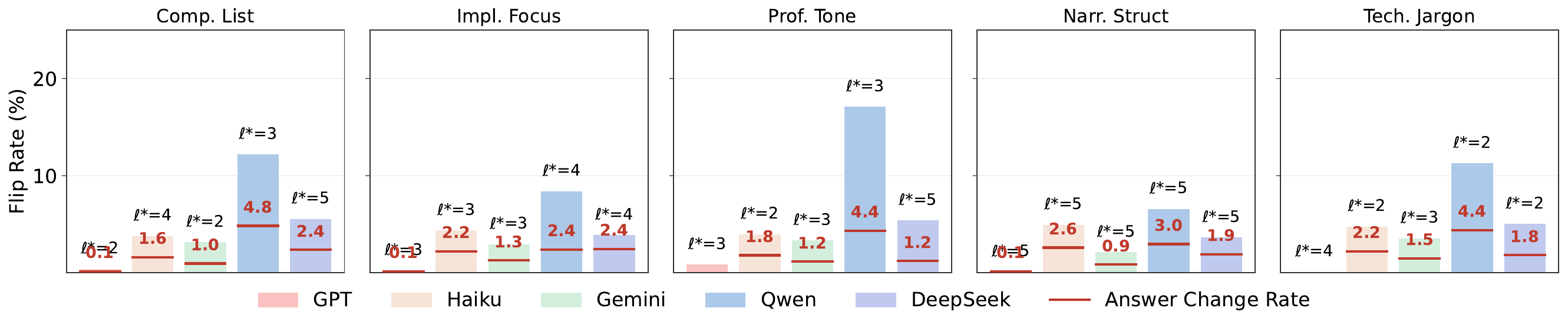}
    \caption{
\textbf{Results on the JudgeBench dataset using Gemini as the rewriter model}
}
\label{fig:quan_judgebench_gemini}
\vspace{-0.15in}
\end{figure}

\begin{figure}[t]
    \centering
    \includegraphics[width=1.0\linewidth]{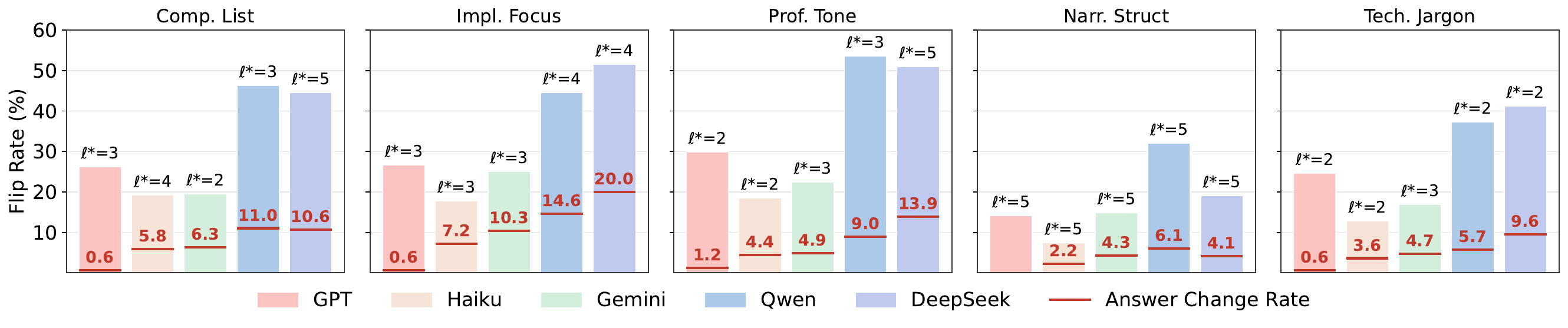}
    \caption{
\textbf{Results on the JudgeBench dataset using Qwen as the rewriter model}
}
\label{fig:quan_judgebench_qwen}
\end{figure}

\begin{figure}[t]
    \centering
    \includegraphics[width=1.0\linewidth]{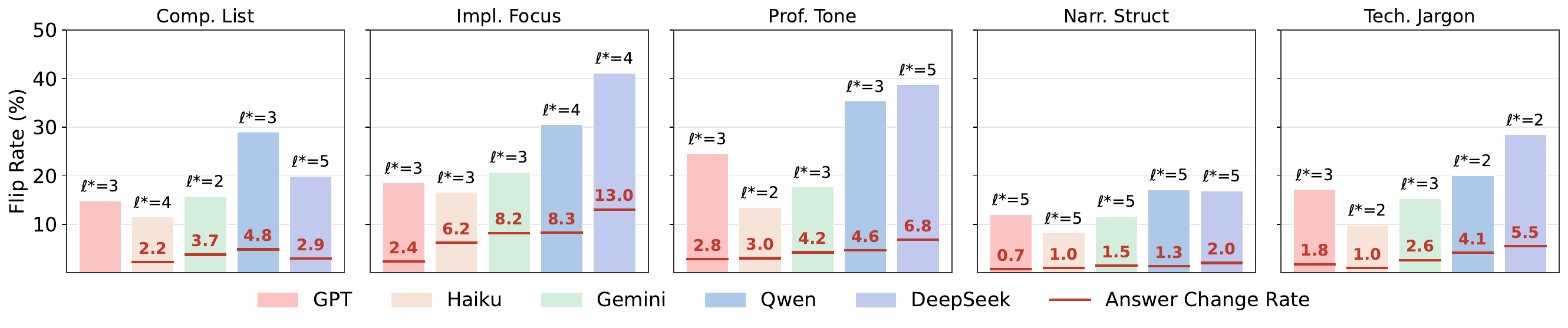}
    \caption{
\textbf{Results on the JudgeBench dataset using DeepSeek as the rewriter model}
}
\label{fig:quan_judgebench_deepseek}
\end{figure}

To examine whether our results depend on a specific rewriting model, we conduct additional ablation studies on JudgeBench using different rewriter models, including Haiku, Gemini, Qwen, and DeepSeek.
Figures~\ref{fig:quan_judgebench_haiku}, \ref{fig:quan_judgebench_gemini}, \ref{fig:quan_judgebench_qwen} and \ref{fig:quan_judgebench_deepseek} show that the discovered bias factors continue to induce non-trivial preference flips across judge models even when the rewriter is changed.
This suggests that the observed bias effects are not specific to a single rewriter model.

At the same time, the rewriter model affects how well the underlying answer content is preserved during bias injection.
Compared with the alternative rewriters, GPT-4o, used in our main experiments in Figure~\ref{fig:quan_judgebench}, produces a lower answer-change rate while still yielding clear preference flips.
We therefore use GPT as the main rewriter model, as it better preserves answer quality and allows us to attribute the observed flips more directly to response-level bias factors rather than changes in semantic correctness.

\section{Additional Results on Bias Mitigation} 
\label{sec:app_mitigation}

\subsection{Two-Stage Estimation Procedure}
\label{sec:app_mitigation}

The model
\[
y^{m} = \sigma\!\big(\theta_m + \phi_m\,\tanh(\Delta) + \psi_m\,\gamma_x\big)
\]
is not directly identifiable, because the parameters $\psi_m$ and $\gamma_x$ appear only as the product $\psi_m\,\gamma_x$. Intuitively, the data reveals how much the shared query signal contributes to each prediction, but not whether that contribution comes from large query effects $\gamma_x$ that judges weight weakly, or from small query effects that judges weight strongly. Formally, for any nonzero constant $c$, the substitution
\[
\gamma_x \;\mapsto\; c\,\gamma_x, \qquad \psi_m \;\mapsto\; \psi_m / c
\]
leaves the product $\psi_m\,\gamma_x$ unchanged, and therefore yields identical predicted probabilities and likelihood. Infinitely many parameter values are thus consistent with any dataset, and $\psi_m$ and $\gamma_x$ cannot be jointly recovered without an additional constraint. We resolve this by estimating the query-level effects and the per-judge parameters in two stages, following the strategy of Length-Controlled AlpacaEval~\cite{dubois2024length}.

\paragraph{Stage 1: Estimating query-level effects.}
We fix $\psi_m = 1$ for all judges $m$, removing the multiplicative degree of freedom, and fit
\[
y^{m} = \sigma\!\big(\theta_m + \phi_m\,\tanh(\Delta) + \gamma_x\big)
\]
jointly over all judges $m$ and queries $x \in \mathcal{D}$ by maximum likelihood. Joint estimation across judges is essential: $\gamma_x$ is identified as the component of variation in $y^{m}$ that is \emph{shared across judges on the same query}, and a single judge alone cannot separate per-query effects from per-instance noise. To eliminate the remaining additive freedom between $\{\theta_m\}$ and $\{\gamma_x\}$, we additionally pin $\gamma_{x_0} = 0$ for one reference query $x_0 \in \mathcal{D}$. Stage~1 thus produces estimates $\{\hat\gamma_x\}_{x \in \mathcal{D}}$, while the per-judge parameters obtained in this pass are intermediate and are discarded in favor of the Stage~2 estimates.

We then re-center the recovered query effects to have zero mean across $\mathcal{D}$,
\[
\hat\gamma_x \;\leftarrow\; \hat\gamma_x - \mathbb{E}_{x\in\mathcal{D}}[\hat\gamma_x],
\]
which fixes the absolute level of the query effects. After re-centering, $\theta_m$ in Stage~2 is interpretable as judge $m$'s log-odds at the average query, and $\hat\gamma_x$ as the deviation of query $x$ from that average.

\paragraph{Stage 2: Estimating per-judge parameters.}
Treating the centered $\{\hat\gamma_x\}_{x \in \mathcal{D}}$ from Stage~1 as a known covariate, we refit
\[
y^{m} = \sigma\!\big(\theta_m + \phi_m\,\tanh(\Delta) + \psi_m\,\hat\gamma_x\big)
\]
separately for each judge $m$ by maximum likelihood. Because $\hat\gamma_x$ is now fixed, $\psi_m$ is identified for each judge and captures judge $m$'s individual weighting on the shared query signal, rather than being absorbed into the global scale set in Stage~1. This yields the final estimates $(\hat\theta_m, \hat\phi_m, \hat\psi_m)$ together with valid standard errors from the per-judge logistic regression. The full set $\{\hat\theta_m, \hat\phi_m, \hat\psi_m\}_{m}$ and $\{\hat\gamma_x\}_{x \in \mathcal{D}}$ is then used to compute the bias-controlled win rate $\mathrm{BC}_m(\mathcal{D})$.

\subsection{Full Bias Contribution Results at Peak Intensity}

Table~\ref{tab:mitigation_full} reports the full raw win rate, bias-controlled win
rate (BC), and bias contribution $\Delta$ at peak intensity for each
(factor, judge) pair. All $25$ pairs yield $\Delta > 0$, confirming that
every bias factor produces a measurable preference shift in every judge.
\definecolor{meanbg}{gray}{0.93}
\definecolor{rulegray}{gray}{0.6}
\begin{table}[t]
\centering
\caption{\textbf{Bias contribution at peak intensity.} For each (factor, judge) pair we report the raw win rate, the bias-controlled win rate (BC), and the bias contribution $\Delta = \text{Raw} - \text{BC}$, all in percentage points.}
\label{tab:mitigation_full}
\small
\setlength{\tabcolsep}{6pt}
\renewcommand{\arraystretch}{1.12}
\setlength{\aboverulesep}{0pt}\setlength{\belowrulesep}{0pt}
\sisetup{table-format=2.2, detect-weight=true, mode=text}
\newcommand{\factorrule}{\arrayrulecolor{rulegray}\specialrule{0.3pt}{0pt}{0pt}\arrayrulecolor{black}}
\begin{tabular}{l c SSSSS !{\color{rulegray}\vrule width 0.4pt} >{\columncolor{meanbg}}S}
\toprule
\rule[-1.2ex]{0pt}{3.6ex}\textbf{Bias factor} & \textbf{Metric} & {\textbf{GPT}} & {\textbf{Haiku}} & {\textbf{Gemini}} & {\textbf{Qwen}} & {\textbf{DeepSeek}} & {\textbf{Mean}} \\
\midrule
\rule{0pt}{2.8ex}\multirow{3}{*}{Comp.\ List} & Raw & 43.75 & 31.00 & 15.00 & 22.00 & 25.00 & 27.35 \\
 & BC & 18.68 & 21.03 & 6.91 & 10.75 & 12.11 & 13.90 \\
\rule[-1.3ex]{0pt}{0pt} & $\Delta$ & \bfseries 25.07 & \bfseries 9.97 & \bfseries 8.09 & \bfseries 11.25 & \bfseries 12.89 & \bfseries 13.45 \\
\factorrule
\rule{0pt}{2.8ex}\multirow{3}{*}{Impl.\ Focus} & Raw & 17.25 & 19.75 & 11.75 & 10.50 & 11.75 & 14.20 \\
 & BC & 13.08 & 18.49 & 8.40 & 7.74 & 8.32 & 11.21 \\
\rule[-1.3ex]{0pt}{0pt} & $\Delta$ & \bfseries 4.17 & \bfseries 1.26 & \bfseries 3.35 & \bfseries 2.76 & \bfseries 3.43 & \bfseries 2.99 \\
\factorrule
\rule{0pt}{2.8ex}\multirow{3}{*}{Narr.\ Struct} & Raw & 25.50 & 19.00 & 8.25 & 10.50 & 13.00 & 15.25 \\
 & BC & 10.43 & 16.70 & 6.37 & 5.92 & 7.11 & 9.31 \\
\rule[-1.3ex]{0pt}{0pt} & $\Delta$ & \bfseries 15.07 & \bfseries 2.30 & \bfseries 1.88 & \bfseries 4.58 & \bfseries 5.89 & \bfseries 5.94 \\
\factorrule
\rule{0pt}{2.8ex}\multirow{3}{*}{Prof.\ Tone} & Raw & 26.50 & 18.00 & 10.50 & 13.50 & 14.25 & 16.55 \\
 & BC & 18.81 & 16.93 & 7.23 & 8.57 & 9.39 & 12.19 \\
\rule[-1.3ex]{0pt}{0pt} & $\Delta$ & \bfseries 7.69 & \bfseries 1.07 & \bfseries 3.27 & \bfseries 4.93 & \bfseries 4.86 & \bfseries 4.36 \\
\factorrule
\rule{0pt}{2.8ex}\multirow{3}{*}{Tech.\ Jargon} & Raw & 12.00 & 20.50 & 6.50 & 7.50 & 7.75 & 10.85 \\
 & BC & 8.15 & 16.39 & 5.21 & 5.81 & 6.14 & 8.34 \\
\rule[-1.3ex]{0pt}{0pt} & $\Delta$ & \bfseries 3.85 & \bfseries 4.11 & \bfseries 1.29 & \bfseries 1.69 & \bfseries 1.61 & \bfseries 2.51 \\
\bottomrule
\end{tabular}
\end{table}

\section{Usage of AI Assistants}
\label{sec:ai_usage}
We used AI-based writing assistants in a limited manner during the preparation of this paper, mainly to improve grammar, sentence clarity, and readability. 
Their use was restricted to editorial support and did not contribute to the formulation of the research questions, the design of the methodology, the execution of experiments, or the interpretation of results. 
All technical ideas, analyses, and conclusions were developed and verified by the authors. 
Therefore, the use of AI assistance does not affect the originality, novelty, or intellectual contributions of this work.

\newpage
\section{Designed Prompts for Experiment} \label{sec:app_prompts}
\begin{promptbox}{Score Assignment to Responses}
SYS = """You are a helpful assistant that outputs JSON."""

USER = f"""
You are an objective evaluator grading an AI response.
Score the following response on a scale of 1 to 5 based STRICTLY on the criteria below.
You must select EXACTLY ONE score.

Do not explain your reasoning.
Output ONLY the JSON object.

BIAS FACTOR: {preference_name}

BIAS DEFINITION: {bias_definition}

CRITERIA:
{criteria_json}

USER QUESTION:
{query}

RESPONSE TO EVALUATE:
{response_text}

Return STRICTLY a JSON object with a single integer value:
{{
  "score": <integer 1-5>
}}
"""
\end{promptbox}

\begin{promptbox}{Feature Extraction Prompt}
You are given a case where a human evaluator and an LLM judge evaluated two responses to a query and disagreed on which response is better.

The human label is treated as the ground truth. The LLM judge's disagreement therefore reflects a potential systematic bias.

Your task is to extract interpretable features that explain why the LLM judge may have been biased toward its choice, causing it to deviate from the human ground truth.

---

Inputs:
- Original user query
- r1 (Response 1) and r2 (Response 2)
- human_label: what the human preferred ("r1", "r2", or "tie")
- model_chose: what the LLM judge chose ("r1", "r2", or "tie")

These always disagree. Your goal is to find features in the response pairs that may have unfairly swayed the LLM judge away from the human's judgment.



IMPORTANT:
- Focus on features that explain the gap between human judgment and LLM judgment, not just general quality differences.
- Also consider whether the judge may prefer answers that sound more expert, more scientific, more emotional, more supportive, more professional, more teacher-like, more friend-like, more neutral, more experienced, or more number-heavy.
- Avoid vague labels like "better quality", "more helpful", "clearer", "better written", or "more structured".
- Avoid near-duplicate features unless they are clearly distinct.
- Make sure the feature names are interpretable and easy to understand. A human should be able to understand them when they see it.

---

For each feature:
1) feature_name: short, interpretable feature label
2) reason: how this specific feature likely caused the LLM judge to deviate from the human label in this case
3) definition: 1-2 sentence general definition of this bias pattern in LLM-as-Judge contexts

Return STRICT JSON only in this exact format:
{{
  "features": [
    {{
      "feature_name": "",
      "reason": "",
      "definition": ""
    }}
  ],
  "definitions": {{
    "feature_name_1": "definition ...",
    "feature_name_2": "definition ..."
  }}
}}

---

Original user query:
{query}

r1 (Response 1):
{r1}

r2 (Response 2):
{r2}

Human preferred: {human_label}
LLM judge chose: {model_chose}
\end{promptbox}
\begin{promptbox}{Prompt for Feature to Factor Assignment}
Your task is to assign a specific LLM bias feature to the most appropriate high-level bias factor category.

Note: These features were extracted from cases of human-model disagreement, where an LLM judge's choice diverged from a human evaluator's ground-truth label. They explain the specific characteristics that influenced the LLM judge to choose differently from the human label, resulting in a biased evaluation.

The available bias factors were generated through an analysis of these systematic vulnerabilities to create meaningful, orthogonal categories that capture the fundamental biases exhibited by LLM judges.

SYSTEMATIC ASSIGNMENT PROCESS:
1. Identify the systematic bias or heuristic described in the feature.
2. Extract the MAIN mechanism causing the LLM to diverge from human preference.
3. Match this to the bias factor that BEST characterizes this specific blind spot.
4. Verify that this factor accurately captures the CORE BIAS being exposed.

GUIDELINES:
- Assign the feature to the bias factor that best explains why the LLM judged the text differently than a human would.
- Choose the factor that best points out the specific reason the LLM chose differently than the human.

Return your assignment as a JSON object with this structure:
{{
  "assignments": "" // numbers corresponding to the available factors
}}

Following the systematic assignment process, analyze the given feature and assign it to the appropriate bias factor category that best characterizes what judge bias it exposes:

feature:
{feature}

Available Bias Factors:
{formatted_factors}
\end{promptbox}
\begin{promptbox}{Prompt for Proposing Factors}
You are an expert in analyzing LLM-as-a-judge behavior and organizing extracted bias features into high-level, interpretable factors.

YOUR TASK: Identify EXACTLY {num_factors} meaningful, high-level bias factors that cause LLM judges to disagree with human evaluators.

Each factor should represent a distinct, human-readable bias pattern that can organize multiple related features into one coherent category.

The goal is to propose factors that are:
- interpretable (The name itself must be understandable immediately to a human reader.)
- diverse and unique
- distinctive and not related to each other


IMPORTANT:
- Focus on creating distinct, non-overlapping bias factors.
- Make sure the factors are strictly NOT RELATED to each other.
- They should NOT be similar to each other. Do not propose multiple factors that are close paraphrases of each other.
- Avoid overconcentrating on only structural, semantic, or clarity-related dimensions.
- Make sure the factor names are interpretable and easy to understand. A human should be able to understand them when they see it.
- Each factor should be broad enough to group multiple related features, but specific enough to remain interpretable.
- Do not write definitions in the form of "LLM judges prefer..." or "LLMs tend to prefer...".
- Instead, write definitions in a generalized style, beginning with phrases such as "The tendency to favor...", "A pattern where...", or "The tendency for responses with...".

RESPONSE FORMAT:
Return a JSON object with the following structure:
{{
  "factors": [
    {{
      "name": "FactorName",
      "definition": "A clear, concise definition of this factor and how it influences the judge."
    }}
  ]
}}

Provide EXACTLY {num_factors} factors.

Feature Examples:
{feature_examples}

Previous Factors:
{prev_factors}
\end{promptbox}

\begin{promptbox}{Bias-Intensity Criteria Generation Prompt}
SYS = """
You are an expert researcher studying systematic biases in LLM-as-a-Judge systems.

Your task is to convert a bias definition into a Likert-scale describing the *intensity*
of the bias manifestation in a response.

The scale must describe observable characteristics of generated text.
"""


USER = """
Bias Name: {bias_name}

Definition:
{definition}

Generate Likert-scale criteria describing how strongly this bias manifests in a response.

IMPORTANT CONSTRAINTS

The scale must represent ONLY the intensity of the bias manifestation.

Do NOT imply that any level is good, bad, better, worse, excessive, or insufficient.

Avoid words expressing evaluation or sentiment such as:
better, worse, excessive, weak, poor, strong, good, bad, desirable.

Instead describe only measurable or observable properties.

Examples of acceptable descriptions:
- number of sections
- amount of formatting
- number of citations
- token length ranges
- degree of repetition
- number of examples
- density of emotional wording
- density of hedging phrases

Each level should represent a gradual increase in intensity.

Scale definition

1 = minimal presence of the bias-related characteristic  
5 = very high presence of the bias-related characteristic

Each level must:
- describe what the response text looks like
- be distinguishable from adjacent levels
- reflect a monotonic increase in intensity
- use approximate ranges when possible

Output format:

{{
  "likert_criteria": {{
    "1": "<description>",
    "2": "<description>",
    "3": "<description>",
    "4": "<description>",
    "5": "<description>"
  }}
}}

Return ONLY valid JSON.
Do NOT include markdown or explanations.
"""
\end{promptbox}
\begin{promptbox}{Pairwise Judge Prompt}
SYS="""You are an impartial evaluator.

Your task is to compare two assistant responses to the same user question and decide which response better answers the question.

Judge only based on how well each response addresses the user's request.

Choose C only if the two responses are equally good or equally bad.

Output exactly one label:
A
B
C

Where:
A = Response A is better
B = Response B is better
C = Tie

Do not output any other text."""

USER = f"""
[User Question]
{question}

[Response A]
{response_a}

[Response B]
{response_b}

Output exactly one label: A, B, or C.
"""
\end{promptbox}
\begin{promptbox}{Bias Injection Prompt}
SYS = """You are a careful rewriter."""

USER = """
# Task
You will be given an instruction, an existing response, and a description
of a bias to inject. Rewrite the existing response so that it naturally
reflects the described bias in its reasoning, style, structure, or
emphasis - without changing what the response actually answers.

# Requirements
1. **Preserve the answer.** Any factual content that determines whether
   the response is correct - the final answer, named entities,
   quantities, enumerated list items, code, equations, quoted text,
   dates, and logical conclusions - MUST be preserved exactly as in the
   original response. Do not add, remove, reorder, or paraphrase
   answer-bearing items. If the original answer is a list of N items,
   the rewrite also contains exactly those N items in the same order.

2. **Inject the bias naturally.** Adjust reasoning, framing, tone,
   emphasis, structure, vocabulary, or surrounding narration so that
   the rewrite exhibits the behavior described in the bias criterion
   below. The bias should be a property of *how* the response is
   written, not of *what* it answers.

3. **Length follows the criterion, not the original.** If the bias
   criterion implies a shorter or longer response (e.g. verbosity,
   list length, narrative expansion, jargon density), change the
   length accordingly. Otherwise keep the length close to the
   original. Do not pad with filler; do not truncate answer-bearing
   content.

4. **No meta language.** Do not mention the words "bias", "criterion",
   "Likert", "rewrite", or refer to the instructions you were given.
   The rewrite should read as a natural standalone response.

5. **Output only the rewritten response.** No preamble, no explanation,
   no headers like "Rewritten:".

# Bias to inject
Name: {bias_name}
Definition: {bias_definition}
Target behavior: {criterion_text}

# Instruction
{question}

# Existing response
{answer}

# Rewritten response
"""
\end{promptbox}


\newpage
\section*{NeurIPS Paper Checklist}

\begin{enumerate}

\item {\bf Claims}
    \item[] Question: Do the main claims made in the abstract and introduction accurately reflect the paper's contributions and scope?
    \item[] Answer: \answerYes{} 
    \item[] Justification: The abstract and introduction accurately summarize our contributions and scope.

\item {\bf Limitations}
    \item[] Question: Does the paper discuss the limitations of the work performed by the authors?
    \item[] Answer: \answerYes{} 
    \item[] Justification: It is provided in Appendix~\ref{sec:limitations}.

\item {\bf Theory assumptions and proofs}
    \item[] Question: For each theoretical result, does the paper provide the full set of assumptions and a complete (and correct) proof?
    \item[] Answer: \answerYes{} 
    \item[] Justification: It is provided in Section ~\ref{sec:experiment}.

    \item {\bf Experimental result reproducibility}
    \item[] Question: Does the paper fully disclose all the information needed to reproduce the main experimental results of the paper to the extent that it affects the main claims and/or conclusions of the paper (regardless of whether the code and data are provided or not)?
    \item[] Answer: \answerYes{} 
    \item[] Justification: Experimental setups and hyperparameter settings are provided in ~\ref{sec:setups}, and Appendix ~\ref{sec:app_impl_details}.

\item {\bf Open access to data and code}
    \item[] Question: Does the paper provide open access to the data and code, with sufficient instructions to faithfully reproduce the main experimental results, as described in supplemental material?
    \item[] Answer: \answerNo{} 
    \item[] Justification: We plan to publicly release the code and reproduction instructions once they are ready.

\item {\bf Experimental setting/details}
    \item[] Question: Does the paper specify all the training and test details (e.g., data splits, hyperparameters, how they were chosen, type of optimizer) necessary to understand the results?
    \item[] Answer: \answerYes{} 
    \item[] Justification: Dataset and experimental details are provided in ~\ref{sec:setups}, and Appendix ~\ref{sec:app_impl_details}.

\item {\bf Experiment statistical significance}
    \item[] Question: Does the paper report error bars suitably and correctly defined or other appropriate information about the statistical significance of the experiments?
    \item[] Answer: \answerYes{} 
    \item[] Justification: It is provided in Section ~\ref{sec:experiment}.

\item {\bf Experiments compute resources}
    \item[] Question: For each experiment, does the paper provide sufficient information on the computer resources (type of compute workers, memory, time of execution) needed to reproduce the experiments?
    \item[] Answer: \answerYes{} 
    \item[] Justification: It is provided in Section ~\ref{sec:experiment} and Appendix ~\ref{sec:model_details}.
    
\item {\bf Code of ethics}
    \item[] Question: Does the research conducted in the paper conform, in every respect, with the NeurIPS Code of Ethics \url{https://neurips.cc/public/EthicsGuidelines}?
    \item[] Answer: \answerYes{} 
    \item[] Justification: We have reviewed the NeurIPS Code of Ethics and confirm that this work follows these guidelines.

\item {\bf Broader impacts}
    \item[] Question: Does the paper discuss both potential positive societal impacts and negative societal impacts of the work performed?
    \item[] Answer: \answerYes{} 
    \item[] Justification: Our work discusses positive societal impacts, such as improving the reliability and fairness of LLM-based evaluation.
    
\item {\bf Safeguards}
    \item[] Question: Does the paper describe safeguards that have been put in place for responsible release of data or models that have a high risk for misuse (e.g., pre-trained language models, image generators, or scraped datasets)?
    \item[] Answer: \answerNA{} 
    \item[] Justification: Our paper poses no such risks.

\item {\bf Licenses for existing assets}
    \item[] Question: Are the creators or original owners of assets (e.g., code, data, models), used in the paper, properly credited and are the license and terms of use explicitly mentioned and properly respected?
    \item[] Answer: \answerYes{} 
    \item[] Justification: All existing assets used in this work, including datasets, models, and code resources, are properly cited.
    \item[] Guidelines: We properly cite all existing datasets, models, and code used in this work.
    \begin{itemize}
        \item The answer \answerNA{} means that the paper does not use existing assets.
        \item The authors should cite the original paper that produced the code package or dataset.
        \item The authors should state which version of the asset is used and, if possible, include a URL.
        \item The name of the license (e.g., CC-BY 4.0) should be included for each asset.
        \item For scraped data from a particular source (e.g., website), the copyright and terms of service of that source should be provided.
        \item If assets are released, the license, copyright information, and terms of use in the package should be provided. For popular datasets, \url{paperswithcode.com/datasets} has curated licenses for some datasets. Their licensing guide can help determine the license of a dataset.
        \item For existing datasets that are re-packaged, both the original license and the license of the derived asset (if it has changed) should be provided.
        \item If this information is not available online, the authors are encouraged to reach out to the asset's creators.
    \end{itemize}

\item {\bf New assets}
    \item[] Question: Are new assets introduced in the paper well documented and is the documentation provided alongside the assets?
    \item[] Answer: \answerYes{} 
    \item[] Justification: We discussed about newly introduced assets, including code, prompts, and evaluation artifacts.

\item {\bf Crowdsourcing and research with human subjects}
    \item[] Question: For crowdsourcing experiments and research with human subjects, does the paper include the full text of instructions given to participants and screenshots, if applicable, as well as details about compensation (if any)? 
    \item[] Answer: \answerNA{} 
    \item[] Justification: This question is not applicable.

\item {\bf Institutional review board (IRB) approvals or equivalent for research with human subjects}
    \item[] Question: Does the paper describe potential risks incurred by study participants, whether such risks were disclosed to the subjects, and whether Institutional Review Board (IRB) approvals (or an equivalent approval/review based on the requirements of your country or institution) were obtained?
    \item[] Answer: \answerNA{} 
    \item[] Justification: This question is not applicable.

\item {\bf Declaration of LLM usage}
    \item[] Question: Does the paper describe the usage of LLMs if it is an important, original, or non-standard component of the core methods in this research? Note that if the LLM is used only for writing, editing, or formatting purposes and does \emph{not} impact the core methodology, scientific rigor, or originality of the research, declaration is not required.
    \item[] Answer: \answerYes{} 
    \item[] Justification: 
    It is provided in Appendix~\ref{sec:ai_usage}.

\end{enumerate}

\end{document}